\documentclass{article}

\usepackage[preprint]{neurips_2022}

\usepackage[utf8]{inputenc} 
\usepackage[T1]{fontenc}    
\usepackage{hyperref}       
\usepackage{url}            
\usepackage{amsfonts}       
\usepackage{nicefrac}       
\usepackage{microtype}      

\usepackage{caption}
\usepackage{subcaption}
\usepackage[table]{xcolor}
\usepackage{graphicx}

\usepackage{amssymb}
\usepackage{latexsym}
\usepackage{amsmath}
\usepackage{amsthm}
\usepackage{eucal}
\usepackage{amstext}
\usepackage{dsfont}
\usepackage{arydshln}

\usepackage{booktabs}
\usepackage{threeparttable}
\usepackage{dashrule}
\usepackage{algorithm,algpseudocode}
\usepackage{pdflscape}

\newcommand{\vect}[1]{\mathbf{#1}}

\usepackage{xr}

\usepackage[capitalize]{cleveref}
\crefname{section}{Section}{Sections}
\Crefname{section}{Section}{Sections}
\crefname{table}{Table}{Tables}
\Crefname{table}{Table}{Tables}
\crefname{figure}{Figure}{Figures}
\Crefname{figure}{Figure}{Figures}

\DeclareMathOperator {\gap}{GAP}

\newcommand{\bftab}{\fontseries{b}\selectfont}

\title{{P}-{NOC}: adversarial training of {CAM} generating networks for robust weakly supervised semantic segmentation priors}

\author{%
  Lucas~David\\ 
  Institute of Computer Science\\
  University of Campinas\\
  Campinas, Brazil\\
  \texttt{lucas.david@ic.unicamp.br} \\
  \And
  Helio~Pedrini \\
  Institute of Computer Science\\
  University of Campinas\\
  Campinas, Brazil\\
  \texttt{helio@ic.unicamp.br} \\
  \And
  Zanoni~Dias \\
  Institute of Computer Science\\
  University of Campinas\\
  Campinas, Brazil\\
  \texttt{zanoni@ic.unicamp.br} \\
}

\begin{document}

\maketitle

\begin{abstract}
  Weakly Supervised Semantic Segmentation (WSSS) techniques explore individual regularization strategies to refine Class Activation Maps (CAMs). In this work, we first analyze complementary WSSS techniques in the literature, their segmentation properties, and the conditions in which they are most effective. {Based on these findings, we devise two new techniques: P-NOC and C²AM-H. In the first, we promote the conjoint training of two adversarial CAM generating networks: the generator, which progressively learns to erase regions containing class-specific features, and a discriminator, which is refined to gradually shift its attention to new class discriminant features. In the latter, we employ the high quality pseudo-segmentation priors produced by P-NOC to guide the learning to saliency information in a weakly supervised fashion.} Finally, we employ both pseudo-segmentation priors and pseudo-saliency proposals in the random walk procedure, resulting in higher quality pseudo-semantic segmentation masks, and competitive results with the state of the art.
\end{abstract}


\section{Introduction}
\label{sec:intro}

Image Segmentation is a paramount component in any autonomous imagery reading system, and also one of its most challenging tasks~\cite{bhanu2012genetic}. Semantic Segmentation consists in correctly associating each pixel of an image to a specific class from a predefined set, and is, to this day, one of the most prominent topics of study in Computer Vision~\cite{mo2022review}, considering its applicability and effectiveness over multiple real-world automation problems from various domains, such as self-driving vehicles~\cite{liu2020importance}, autonomous environment surveillance and violence detection~\cite{gelana2019firearm,hernandez2012detecting}, vegetation detection in satellite imaging~\cite{zhan2020vegetation} and medical imagery analysis~\cite{lalonde2021capsules,xie2020skin}.

Modern approaches based on Convolutional Networks stand out by consistently outscoring classic techniques with great effectiveness across different areas and datasets~\cite{chen2020research}. However, these require massive amounts of pixel-level annotated information, obtained through extensive human supervision. Given time and cost constraints, these solutions remain inaccessible to many.

To circumvent the aforementioned limitations, researchers often recur to Weakly Supervised Semantic Segmentation (WSSS)~\cite{ouassit2022brief}, where ``weakly'' refers to partially supervised information, or lack thereof. Recent work investigated deriving segmentation maps from saliency maps, bounding boxes, scribes and points, and even image-level annotations~\cite{zhang2020survey}.

A typical solution to a WSSS problem, comprising only image-level annotations, consists of training a classification model and devise pseudo-semantic segmentation proposals from localization cues extracted from visual explaining methods, such as Class Activation Mapping (CAM)~\cite{Zhou_2016_CVPR} and Grad-CAM~\cite{Selvaraju2017gradcam8237336}. Moreover, WSSS techniques often exploit complex regularization strategies that instigate the development of properties that are useful to the segmentation task, such as prediction completeness~\cite{vilone2020explainable,puzzle9506058}, fidelity to semantic boundaries~\cite{li2018tell,wei2017object,occse9711138}, awareness of saliency and contextual information~\cite{lee2021reducingbottleneck,xie2022c2am}, and robustness against noise~\cite{muller2019labelsmoothing,yun2019cutmix}.

While steady improvement is made by state-of-the-art WSSS techniques, many studies fail to report the conditions in which their techniques are most effective, the properties most prominently affected by them, and their behavior when conjointly employed with complementary regularization methods.
Such analysis remains nevertheless paramount for a better understanding of WSSS problems, and for future research concerned with extracting the maximum effectiveness for a given problem.

In this work, we first analyze prominent techniques from WSSS literature. We evaluate their individual performance, and report their capabilities for different groups of classes, considering properties such as occurrence frequency, average size of objects, and co-occurrence rate. We find that these techniques are complementary, as illustrated in \cref{fig:comp-ra-oc-p}: each instigating distinct properties that are better suited for specific contexts and groups within the WSSS task at hand.
With this in mind, we establish a strong \textit{baseline} strategy comprising the conjoint training of the aforementioned techniques.
Finally, we propose three strategies that attempt to instigate multiple properties related to the segmentation task at once, thus creating a solution that performs well in most scenarios.

The main contributions of our work are:

\begin{figure}[!t]
  \centering
  \includegraphics[width=\linewidth]{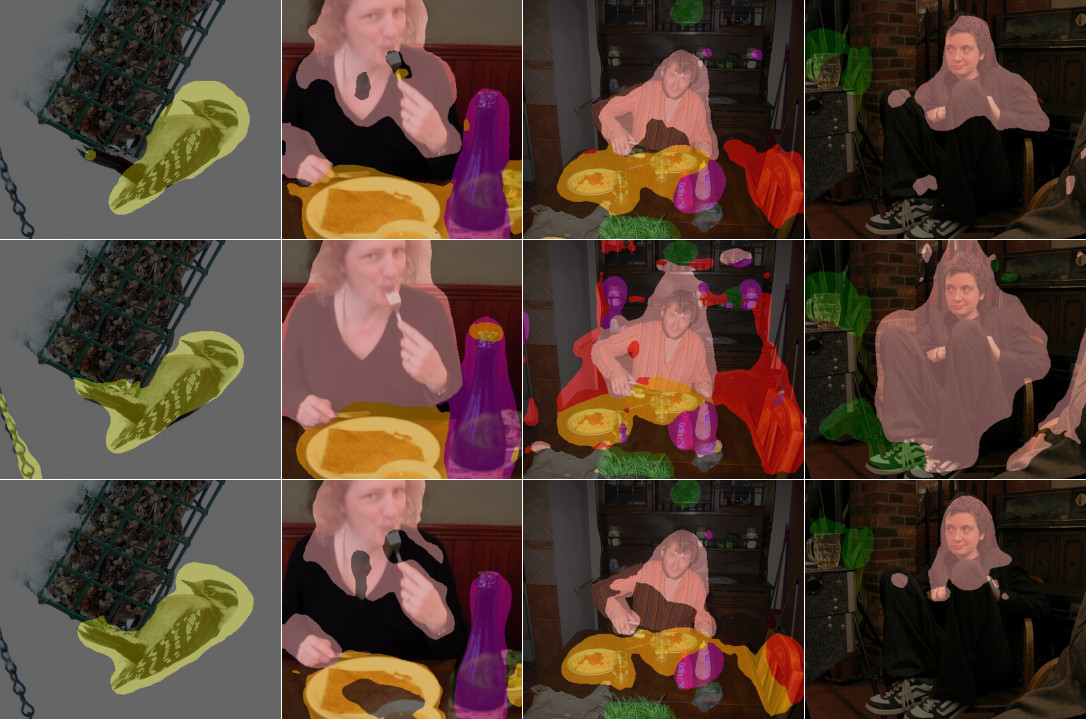}
  \caption{Semantic segmentation priors devised by different WSSS techniques. From top to bottom: (a) CAM, (b) OC-CSE, and (c) Puzzle-CAM.}
  \label{fig:comp-ra-oc-p}
\end{figure}

\begin{enumerate}
    \item \textbf{NOC-CSE:} We propose a new Class-Specific Adversarial Erasing strategy comprising the adversarial training between the generator and a second (not-so-ordinary) network (discriminator): the generator must learn class-specific masks that erase all discriminant features presented to the pretrained discriminator.
    In turn, the discriminator must learn new class-discriminative features.
    Our setup explores the features space more effectively,
    resulting in the segmentation priors with higher quality.
    \item \textbf{C²AM-H:} We propose an extension of the unsupervised saliency detector C²AM~\cite{xie2022c2am} that incorporates ``hints'' of salient regions learned in a weakly supervised setting.
    \item \textbf{Salient affinity labels:} We propose an alternative to the well-established affinity maps (employed in the training of the Affinity Network~\cite{ahn2018learning}) that leverages the learned pseudo-saliency masks to determine reliable regions in the priors, leading to more accurate affinity measurements, and, thus, refined pseudo-semantic segmentation maps with superior quality.
\end{enumerate}

Experimental results suggest that our approach present competitive effectiveness with state-of-the-art literature, even when compared to methods employing higher processing (e.g., attention-based models, Transformers) and loosen data usage constraints (e.g., external saliency information, pre-training over large semantic segmentation sets), over both Pascal VOC 2012~\cite{everingham2015pascal} and MS COCO 2014~\cite{lin2014microsoft} datasets.

The remaining of this work is organized as follows. %
\cref{sec:rw} provides a review of the WSSS literature, describing the most important aspects of semantic segmentation in a weakly supervised setup, and listing previous related research. %
{Additionally, we discuss the most prominent techniques, considering their strengths and shortcomings for different groups and contexts, and further 
detail the fundamentals necessary for the understanding of this work.} %
We describe our approach for improving the quality of semantic segmentation priors in~\cref{sec:method}, our solution for Weakly Supervised Saliency Detection in~\cref{sec:method:wssd}, and our strategy for leveraging pseudo-saliency masks when devising affinity labels in~\cref{sec:method:sal_aff_labels}. %
We describe the experimental settings employed in~\cref{sec:setup}, and report and discuss our results in~\cref{sec:results}. Finally, we summarize our main findings and list directions for future work in~\cref{sec:conclusions}.

\section{Related Work and {Fundamentals of WSSS}}
\label{sec:rw}

In this section, we describe the aspects commonly considered in the weakly supervised literature, and introduce notable techniques previously proposed to approach them.

\subsection{Weakly Supervised Learning}
Multiple WSSS techniques have been devised to approach problems that lack complete supervision, inducing higher quality solutions (measured in \textit{mean Intersection over Union}, or mIoU).
Among many, local attention-based methods~\cite{puzzle9506058,wang2020self,fan2020learning}, Adversarial Erasing (AE) frameworks~\cite{li2018tell,wei2017object,occse9711138}, and saliency-guided methods~\cite{xie2022c2am,fan2020learning,lee2021reducing,lee2021railroad,jiang2022l2g} stand out, entailing notable improvements in the segmentation capabilities across different datasets and domains.

Concomitantly, refinement strategies have also been proposed to further improve the quality of pseudo-segmentation priors.
These strategies exploit pixel-wise similarity and uncertainty in the pseudo-maps to refine them at a pixel-level.
Two methods are frequently employed in the literature: Fully Connected Conditional Random Fields (CRFs)~\cite{krahenbuhl2011efficient}, in which label consistency is reinforced by exploiting visual similarity and smoothness in pixel neighborhood; and Pixel Affinity~\cite{ahn2018learning,ahn2019weakly}, where priors are refined by \textit{random walking}, guided by affinity between pixels.

Despite greatly improving the precision of the pseudo-segmentation maps, refinement methods are strongly influenced by the quality of the seeds.
Thus, authors have studied forms to devise more reliable priors by encouraging the emergence of useful segmentation properties, such as prediction completeness~\cite{vilone2020explainable}, sensitivity to semantic boundaries~\cite{occse9711138}, awareness of saliency and contextual information~\cite{xie2022c2am}, and robustness against noise~\cite{muller2019labelsmoothing,yun2019cutmix}.

\subsection{Aspects Commonly Explored in Weakly Supervised Semantic Segmentation Literature}

\paragraph{Prediction Completeness}
\textit{Prediction completeness} refers to the quality of an explanation (such as CAM), and its capacity to completely describe a prediction or concept~\cite{vilone2020explainable}. In the context of WSSS, it often describes the level of \textit{object coverage} observed in the pseudo-segmentation masks devised.
Attention to local features is paramount to reinforce \textit{prediction completeness}, and it is encouraged by Puzzle-CAM~\cite{puzzle9506058} through the separation of the input image in four equal pieces, forwarding them through a model $f$ and reconstructing the output activation signal by merging the pieces of the puzzle together, which results in an information stream highly focused on local details.
This ``local'' stream is used to regularize the ``main'' Class Activation Maps (CAMs) produced by the model $f$, while it is concomitantly trained to predict multi-label class occurrence from both main and local streams, resulting in more complete and homogeneous activation over all parts of the objects of interest.

\paragraph{Sensitivity to Semantic Boundaries}
Besides \textit{completeness}, another important trait in semantic segmentation is the capacity to distinguish the semantic frontiers of objects associated to different classes.
Adversarial Erasing (AE) methods aim to devise better CAM generating networks by erasing the most discriminative regions during training, forcing the training model to account for the remaining (ignored) features~\cite{li2018tell}.
Class-Specific Erasing (CSE)~\cite{occse9711138}, for instance, performs AE through an assisted training setup between two models ($f$ and $oc$), in which the first must learn maps that erase all features associated with a specific class, hence reducing its detection by the $oc$.
The CAMs learned in this setup become sufficiently accurate to insulate objects of distinct classes, maintaining coarse fidelity to their semantic boundaries.

\paragraph{Robustness Against Noise} Given the noisy nature of the information produced in weakly supervised environments, many authors devised ways of mitigating low-quality data by estimating and ignoring uncertain regions~\cite{li2022uncertainty}, or through augmentation and regularization techniques employed to reduce the impact of mislabeled pixels to the training process, such as CutMix~\cite{yun2019cutmix}, RandAugmentation~\cite{cubuk2020randaugment}, and label smoothing~\cite{muller2019labelsmoothing}.

\paragraph{Awareness of Saliency and Contextual Information}
The utilization of saliency information as complementary information is also unequivocally advantageous for the solution of WSSS tasks, and it is studied by various WSSS techniques~\cite{lee2021reducingbottleneck,lee2021railroad,jiang2022l2g,chaudhry2017discovering,oh8100018exploitingsal,sun2019saliencyguided}. Saliency information can be used to provide additional guidance during the training process of CAM-proposing networks, or simply as saliency maps that, when concatenated with CAMs, result in more accurate pseudo-semantic segmentation masks.
To that end, Xie et al.~proposed C²AM~\cite{xie2022c2am}, an unsupervised technique that employs contrastive learning to find a bi-partition of the spatial field that separates the salient objects from the background. It does so by extracting both low and high level features from the input image using a pretrained feature extracting network, and feeding them to a \textit{disentangling} branch that predicts the probability of each feature belonging to the first partition. Two sets of features (representing background and foreground regions) are extracted, and training ensues by optimizing the model to approximate the representations of the most visually similar patches, while increasing the distance between the foreground and background representations.


\subsection{Complementarity Between WSSS Techniques}
\label{sec:rw:analysis}

Given the complex nature of WSSS, in which one must explore underlying cues on data, solutions are often dependent on multiple criteria being met.
A single solution is unlikely to provide optimal results for all scenarios. This is illustrated in~\cref{fig:comp-ra-oc-p}: (i) naive models may be insufficient to cover all portions of large objects, (ii) while methods that strive for higher \textit{prediction completeness} (e.g., Puzzle-CAM) may produce superior priors for samples containing a single object, (iii) while inadvertently confusing objects of different classes in cluttered scenes.
Similarly, (iv) adversarial strategies (e.g., OC-CSE) can better segment objects with complex boundary shapes, at the risk of over-expanding to visually similar regions.

These findings are also observable in \cref{tbl:intro-wsss-detailed}: (i) CAM significantly underperforms for classes associated with \textit{large} objects ($59.3\%$) and \textit{singleton} classes ($52.4\%$).
On the other hand, (ii) Puzzle-CAM produces the best mIoU for \textit{singleton} classes ($69.3\%$), and (iii) the worst result for classes that frequently co-occur in a \textit{room} ($41.3\%$). Objects in \textit{room} were best segmented by vanilla CAM ($44.6\%$).
Finally, (iv) OC-CSE produces the best results for mid-sized objects ($59.2\%$) and class \textit{person} ($63.8\%$) --- often co-occurring with other classes ---, and balanced scores for the remaining groups.

By correlating the class-specific mIoU values with the properties of that class, we observe a strong linear association between the average relative size (\%S) and the effectiveness of predictions by CAM ($70.3\%$).
The \textit{size} also correlates (with lower intensity) with the remaining techniques, although scores from Puzzle-CAM are more strongly associated with class co-occurrence rate ($-71.0\%$) and average of label set cardinality ($-75.8\%$).
Lastly, OC-CSE display average results once again, further indicating its balanced performance over different class groups.

\begin{table*}[!htb]
\centering
\caption{Scores (in mIoU) observed for class groups in Pascal VOC 2012 \textit{training} set, considering segmentation priors generated by CAM, OC-CSE (CSE), and Puzzle-CAM (P). Class-specific properties, such as label occurrence (O), average relative size (\%S), class co-occurrence rate (\%C) and average of label set cardinality (L) were averaged and listed for inspection purposes.}
\label{tbl:intro-wsss-detailed}
\small
\setlength{\tabcolsep}{1.9mm}
\begin{tabular}{@{}llrrrrrrr@{}}
\toprule
\textbf{\#} & \textbf{Classes} & \textbf{O} & \textbf{\%S} & \textbf{\%C} & \textbf{L} & \textbf{CAM} & \textbf{CSE} & \textbf{P} \\
\midrule
\multicolumn{2}{l}{\textbf{Group}} & \multicolumn{4}{r}{} \\
\rowcolor{gray!10}
Overall & \textit{all} & 108.6 & 19.8\% & 51.2\% & 1.8 & 53.9 & 56.0 & \bftab{61.0} \\
Background & - & - & 69.5\% & - & - & 81.0 & 82.2 & \bftab{86.0} \\
\midrule
\multicolumn{2}{l}{\textbf{Occurrence}} & \multicolumn{4}{r}{} \\
Low freq. & bike boat bus cow horse sheep & 69.3 & 17.6\% & 42.7\% & 1.6 & 51.9 & 58.0 & \bftab{64.5} \\
Mid freq. & a.plane bottle d.table m.bike & 85.0 & 17.3\% & 57.6\% & 1.9 & 52.0 & 50.5 & \bftab{52.9} \\
          & p.plant sofa train tv \\
High freq. & bird car cat chair dog person & 179.2 & 16.9\% & 51.3\% & 1.7 & 53.8 & 57.0 & \bftab{64.0} \\
\midrule
\multicolumn{2}{l}{\textbf{Size}} & \multicolumn{4}{r}{} \\
Small & a.plane bike bird boat & 93.3 & 10.3\% & 50.1\% & 1.8 & 42.4 & 43.2 & \bftab{51.7} \\
      & bottle chair p.plant \\
Mid & car cow person tv & 179.5 & 16.1\% & 59.9\% & 1.9 & 55.0 & \bftab{59.2} & 57.0 \\
Large & bus cat d.table dog horse & 88.9 & 23.2\% & 48.3\% & 1.7 & 59.3 & 61.6 & \bftab{67.1} \\
      & m.bike sheep sofa train \\
\midrule
\multicolumn{2}{l}{\textbf{Group}} & \multicolumn{4}{r}{} \\
Singleton & a.plane bird boat sheep train & 87.4 & 18.2\% & 21.1\% & 1.3 & 52.4 & 58.2 & \bftab{69.3} \\
Person & person & 442.0 & 15.2\% & 83.0\% & 2.1 & 63.6 & \bftab{63.8} & 50.3 \\
Animal & dog horse & 94.5 & 19.5\% & 42.5\% & 1.5 & 58.6 & 65.0 & \bftab{74.9} \\
Room & bottle chair d.table p.plant sofa tv & 96.0 & 15.2\% & 77.0\% & 2.3 & \bftab{44.6} & 43.4 & 41.3 \\
Traffic & bike bus car m.bike & 88.0 & 18.3\% & 61.7\% & 1.9 & 58.7 & 58.0 & \bftab{65.1} \\
\midrule
\multicolumn{2}{l}{\textbf{Correlation}} & \multicolumn{4}{r}{} \\
$\rho$ O   &  & 100.0 &  -6.5 & 30.0  & 20.4  &  19.9 &  10.4 & -12.8 \\
$\rho$ \%S &  &  -6.5 & 100.0 & -18.0 & -23.4 &  70.3 &  51.5 &  48.4 \\
$\rho$ \%C &  &  30.0 & -18.0 & 100.0 & 97.6  & -22.9 & -38.4 & -71.0 \\
$\rho$ L   &  &  20.4 & -23.4 & 97.6  & 100.0 & -32.9 & -44.5 & -75.8 \\
\bottomrule
\end{tabular}
\end{table*}

Going in a different direction, leveraging saliency information (such as in C²AM~\cite{xie2022c2am}) can also drastically improve segmentation effectiveness.
Saliency-based techniques proposed in WSSS often rely on saliency information devised from pretrained saliency detectors, neglecting the underlying risk of data leakage, when these detectors have been pretrained over datasets containing elements (samples, classes, context) that intersect the current segmentation task, and not fine-tuning the detection model to the task at hand.
Conversely, C²AM~\cite{xie2022c2am} provides a simple solution to these problems by employing an unsupervised contrastive approach to learn a saliency discrimination function. Once it is trained, C²AM can be used to estimate saliency maps for each individual image, which can be concatenated to its CAM, offering pixel-level saliency information, and leading to a more accurate pseudo-semantic segmentation priors.

Notwithstanding its notable improvements, C²AM also has drawbacks: (i) relying solely on feature similarity, it is highly influenced by the pretrained feature extractor used.
Furthermore, (ii) no anchor holds the bi-partition function. As a result, training multiple times can inadvertently inverse the direction of the decision function, resulting in models that associate a same object to low and high saliency values, respectively.
Lastly, (iii) the available image-level annotations are not used, possibly resulting in data being under-represented during the training process.

\subsection{Preliminaries}
\label{sec:preliminaries}

In this section, we lay the foundations for our work. We formally define CAM, describe its extraction process, and detail the involved WSSS techniques previously established in the literature that are important for the understanding of this work.

\subsubsection{Class Activation Mapping (CAM)}

    Let $\mathcal{D} = \{(\vect{x}_i, \vect y_i)\}_{i=1}^{N}$ be a training dataset, where $\vect{x}_i$ is the $i$-th sample image in the set, $\vect y_i = [y_i^1, \dots, y_i^{|C|}]$ is the one-hot class label vector indicating which classes are present in image $\vect{x}_i$, and $C$ is the set of all classes.
    Following previous works~\cite{puzzle9506058,occse9711138,ahn2018learning}, we employ a Fully Convolutional Neural Network (CNN) as our classifying model.
    That is, a convolutional feature extractor, followed by a $1\times 1$ convolution layer with $C$ output channels, and by a Global Average Pooling (GAP) layer. Hence, the prediction of the classifier $f$ for the image $\vect{x}_i$ is defined as:
    \begin{equation}
        p_i = \sigma(\gap(f(\vect{x}_i)))
    \end{equation}
    \noindent where \textit{GAP} is the \textit{Global Average Pooling} operation, and $\sigma$ is the \textit{sigmoid} function.
    
    The CAM~\cite{Zhou_2016_CVPR} of a class $c$ is represented by $\vect A_i^c\in\mathbb{R}^{HW}$, and it is obtained by forwarding image $\vect{x}_i$ through the classifier $f$. That is, $\vect A_i^c = f^c(\vect{x}_i)$.
    Finally, a normalization function $\psi: \mathbb{R}\to [0, 1]$ is used so that the CAM can subsequently serve as an erasure mask:
    \begin{equation}
        \psi(\vect A^c) = \frac{\text{ReLU}(\vect{A}^c)}{\max_{hw \in HW}\text{ReLU}(\vect{A}^{hwc})}
    \end{equation}

    Following previous works~\cite{puzzle9506058,occse9711138,ahn2018learning}, we filter the activation signal using the \textit{ReLU} function, thus focusing only on regions that positively contribute to the classification of the class of interest~\cite{Selvaraju2017gradcam8237336}.

\subsubsection{Puzzle-CAM}

    Given a classifying network $f$, we extract $\vect{A}_i^c = f^c(\vect{x}_i) \in \mathbb{R}^{HW}$, the CAM for class $c$ from image $\vect{x}_i$; and ${A^\text{re}}^c_i = \text{merge}(f^c(\text{tile}(\vect{x}_i)))$, the reconstruction of the ``puzzle'' CAM pieces~\cite{puzzle9506058}, which were produced by separating $\vect{x}_i$ into four quadrants (tiles) and forwarding them individually through $f$).
    In these conditions, Puzzle-CAM~\cite{puzzle9506058} is trained by optimizing the following objective functions:
    \begin{equation}\label{eq:p-loss}\begin{split}
        \mathcal{L}_\text{P} &= \mathcal{L}_\text{cls} + \mathcal{L}_\text{re-cls} + \mathcal{L}_\text{re} \\
            &= \ell_\text{cls}(\vect{p}_i, \vect{y}_i)
            + \ell_\text{cls}(\vect{p}^\text{re}_i, \vect{y}_i)
            + \lambda_\text{re}\|\vect{A}_i - \vect{A}^\text{re}_i\|_1
    \end{split}
    \end{equation}

\subsubsection{Ordinary Classifier: Class-specific Adversarial Erasing Framework (OC-CSE)}

    Given $f$ and $oc$, two architecture-sharing CNNs, we extract $\vect{A}_i^c = f^c(\vect{x}_i) \in \mathbb{R}^{HW}$ (the CAM for class $c$ from image $\vect{x}_i$).
    A class --- namely $r$ --- is uniformly sampled from the set of classes associated to $\vect x_i$ (s.t.~$y_i^r = 1$), and its ``back-propagable''~\cite{occse9711138} erasure mask is computed based on its CAM:\begin{equation}
        \bar{\vect{A}}_i^r = 1 -\psi(\vect{A}_i^r)
    \end{equation}
    
    Here, $\bar{\vect{A}}_i^r(h,w)$ represents the probability of pixel $(h,w)$ to \textbf{not} contain parts of objects of class $r$, which is used to mask objects of class $r$ in the original image. The masked image is, in turn, forwarded onto the ordinary classifier \textit{oc}:
    \begin{equation}
        {\vect{A}^\text{oc}}^c_i = \text{oc}^c(\vect{x}_i\circ (1-\psi(\vect{A}_i^r)))
    \end{equation}
    
    From the spatial maps, we obtain the estimated probability of the original and masked input images containing objects of class $c$: $p_i^c(\vect{A}) = \sigma(\gap(\vect{A}_i^c))$, and ${p^\text{oc}}^c_i=\sigma(\gap({\vect{A}^\text{oc}}^c_i))$.
    OC-CSE~\cite{occse9711138} is then trained with the following objective functions in mind:
    \begin{equation}\label{eq:occse-loss}\begin{split}
        \mathcal{L}_\text{OC-CSE} &= \mathcal{L}_\text{cls} + \mathcal{L}_\text{cse} \\
        &= \ell_\text{cls}(\vect{p}_i, \vect{y}_i)
        + \lambda_\text{cse}\ell_\text{cls}(\vect{p}_i^\text{oc}, \vect{y}_i\setminus\{r\})
    \end{split}
    \end{equation}

\subsubsection{Contrastive Learning for Class-agnostic Activation Map (C²AM)}

Let $f$ be a feature extractor s.t.~$f(\vect{x}_i) = \vect{A}_i^{hw}$, and $\varphi:\mathbb{R}^{K}\to[0,1]$ be a ``disentangling'' function, mapping each region in the embedded spatial signal $\vect{A}_i^{hw}$ to the probability value $p_i^{hw}$ of said region belonging to the first partition.
Moreover, let $\vect{v}_i^f = \varphi(\vect{A}_i) \circ \vect{A}_i$ and $\vect{v}_i^b = (1-\varphi(\vect{A}_i)) \circ \vect{A}_i$ be two extracted feature vectors, representing the spatial foreground (\textit{fg}) and background (\textit{bg}) features, respectively.

Considering a batch of $n$ images $\mathcal{B} = \{\vect{x}_b, \vect{x}_{b+1}, \ldots, \vect{x}_{b+n-1}\}$, three cosine similarity matrices are calculated: (a) the \textit{fg} features ($\vect{S}^f = [s^f_{ij}]_{1\leq i,j \leq n}$), (b) the \textit{bg} features ($\vect{S}^b = [s^b_{ij}]_{1\leq i,j \leq n}$); and (c) between the \textit{fg} and \textit{bg} ($\vect{S} = [s^\text{neg}_{ij}]_{1\leq i,j \leq n}$) features.

In these conditions, C²AM~\cite{xie2022c2am} is defined as the optimization of the following objectives:
\begin{equation}
\begin{split}
    \mathcal{L}^\mathcal{B}_\text{C²AM} &= \mathcal{L}^\mathcal{B}_\text{pos-f} + \mathcal{L}^\mathcal{B}_\text{pos-b}
    + \mathcal{L}^\mathcal{B}_\text{neg} \\
    &= \frac{1}{n(n-1)}\sum_i^n\sum_j^n \mathds{1}_{[i\ne j]}(w^f_{ij}\log s^f_{ij}) \\
    &+ \frac{1}{n(n-1)}\sum_i^n\sum_j^n \mathds{1}_{[i\ne j]}(w^b_{ij}\log s^b_{ij}) \\
    &- \frac{1}{n^2}\sum_i\sum_j\log (1-s^\text{neg}_{ij})
\end{split}
\label{eq:ccam-loss}
\end{equation}
\noindent where $w^f_{ij} = e^{-\alpha\text{rank}(s^f_{ij})}$ and $w^b_{ij} = e^{-\alpha\text{rank}(s^b_{ij})}$ are weight factors (for \textit{fg} and \textit{bg} features, respectively) exponentially proportional to the similarity rank between the regions $(i, j)$, considering all possible feature pairs extracted from images in $\mathcal{B}$.

\section{Obtaining Robust Semantic Segmentation Priors}
\label{sec:method}
In this section, we detail our approach to obtain robust semantic segmentation priors.

\subsection{Combining Complementary Strategies for Robustness (Baseline)}
\label{sec:method:poc}

We are drawn to investigate whether the combination of complementary strategies could mitigate their individual shortcomings.
As a baseline for our investigation study, we consider the combination of Puzzle-CAM~\cite{puzzle9506058}, OC-CSE~\cite{occse9711138}, and label smoothing~\cite{muller2019labelsmoothing} to reinforce the best properties in each approach: high \textit{completeness} (to cover all class-specific regions), high sensitivity to semantic boundaries and contours, and robustness against labeling noise.

As described in \cref{sec:preliminaries}, we obtain the estimated probability of $\vect{x}_i$ containing objects of class $c$ from the spatial maps produced by $f$ and $oc$: $p_i^c(\vect{A}) = \sigma(\gap(\vect{A}_i^c))$; ${p^\text{re}}^c_i=\sigma(\gap({\vect{A}^\text{re}}^c_i))$; and ${p^\text{oc}}^c_i=\sigma(\gap({\vect{A}^\text{oc}}^c_i))$.
From them, we define {P-OC} as the optimization of the following objective functions:
\begin{equation}\label{eq:poc-loss}\begin{split}
    \mathcal{L}_\text{P-OC} &= \mathcal{L}_\text{P} + \mathcal{L}_\text{cse} \\
        &= (\mathcal{L}_\text{cls} + \mathcal{L}_\text{re-cls} + \mathcal{L}_\text{re}) + \mathcal{L}_\text{cse} \\
        &= \ell_\text{cls}(\vect{p}_i, \vect{y}_i) + \ell_\text{cls}(\vect{p}^\text{re}_i, \vect{y}_i)
        + \lambda_\text{re}\|\vect{A}_i - \vect{A}^\text{re}_i\|_1 
        + \lambda_\text{cse}\ell_\text{cls}(\vect{p}_i^\text{oc}, \vect{y}_i\setminus\{r\})
\end{split}
\end{equation}
\noindent where $\ell_\text{cls}$ is the multi-label soft margin loss, and $\|\cdot\|_1$ the mean absolute error loss.

\subsection{Adversarial Learning of CAM Generating Networks}
\label{sec:method:pnoc}
The Class-Specific Erasing (CSE) framework~\cite{occse9711138} has shown great promise among the AE methods, noticeably improving WSSS score compared to previous class-agnostic strategies.
When employing an ordinary classifier ($oc$), remaining regions (not erased) associated with a specific class can be quantified, allowing us to better modulate training (paramount to reduce the chronic problem of over-erasure of AE methods~\cite{li2018tell}).

\begin{figure*}[!t]
    \centering
    \begin{subfigure}[t]{\linewidth}
      \includegraphics[width=\linewidth]{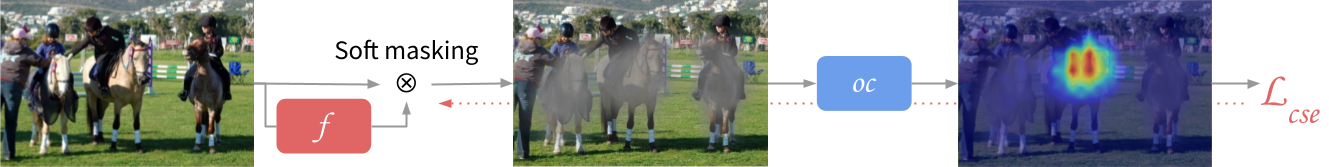}
      \caption{OC-CSE}
    \end{subfigure}
    {\color{gray}\hdashrule[0.5ex][c]{\linewidth}{1pt}{2pt}}
    \begin{subfigure}[t]{\linewidth}
      \includegraphics[width=\linewidth]{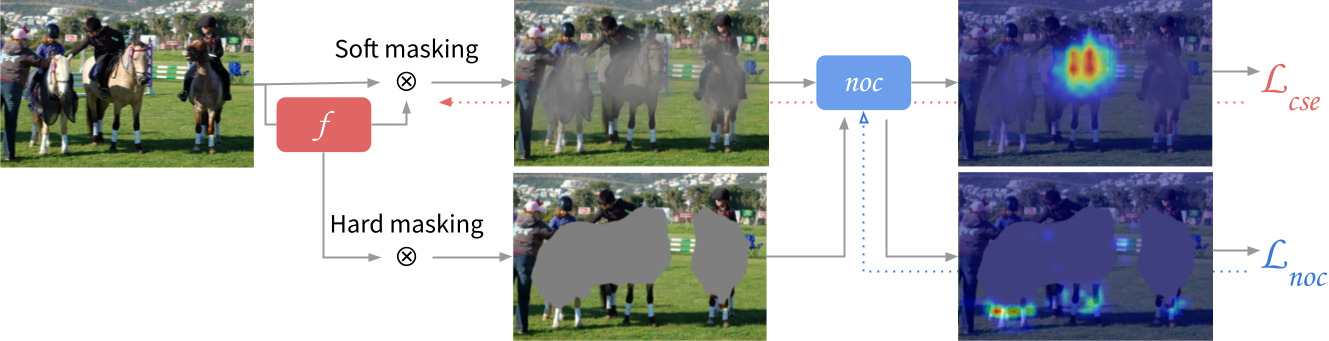}
      \caption{NOC-CSE}
    \end{subfigure}
    \caption{Diagram of our proposed adversarial learning scheme, NOC-CSE.
    (a) OC-CSE: class-specific regions found by $f$ are erased (CSE) with the supervision of a fixed ordinary classifier ($oc$). After many training iterations, core regions associated with a class are saturated, rendering the $oc$ redundant.
    (b) NOC-CSE: $f$ learns class-specific regions with the assistance of a not-so-ordinary classifier ($noc$), which is gradually trained to associate secondary (and yet) discriminant regions to the class of interest.}
    \label{fig:diagram-oc-noc}
\end{figure*}

However, it is unlikely that the $oc$ can provide useful information regarding the remaining uncovered regions once considerable erasure is performed, as (a) the soft masking from a CAM (devised from a non-saturating linear function) can never fully erase the primary features, and (b) the $oc$ has \textbf{only} been trained over whole images, in which marginal regions could be easily ignored.
This behavior is best observed in~\cref{fig:diagram-oc-noc}: after many training iterations under the CSE framework, the $oc$ remains focused on core regions of the class \textit{horse}, notwithstanding these have been mostly erased. Once the input is hard masked, completely removing these regions, the $oc$ can focus on secondary regions (the horses' legs), and thus provide better information about which regions are ignored by $f$.

\paragraph{Method}
To overcome the aforementioned limitation, we propose to extend the CSE framework into a fully adversarial training one (namely NOC-CSE). In it, the generator $f$ must learn class activation maps that erase all regions that contribute to the classification of class $r$, while a second network $noc$ (a ``not-so-ordinary'' classifier) must discriminate the remaining $r$-specific regions neglected by $f$, while learning new features that characterize class $r$, currently ignored by both.
In summary, NOC-CSE is trained by alternatively optimizing two objectives:
\begin{align}
    \mathcal{L}_\text{cse} &= \mathbb{E}_{(\vect{x}_i,\vect{y}_i)\sim\mathcal{D},r\sim \vect{y}_i}[\lambda_\text{cse}\ell_\text{cls}(\vect{p}^\text{oc}, \vect{y}_i\setminus\{r\})] \label{eq:pnoc:f} \\
    \mathcal{L}_\text{noc} &= \mathbb{E}_{(\vect{x}_i,\vect{y}_i)\sim\mathcal{D},r\sim \vect{y}_i}[\lambda_\text{noc}\ell_\text{cls}(\vect{p}_i^\text{noc}, \vect{y}_i)] \label{eq:pnoc:noc}
\end{align}
\noindent where $\vect{p}_i^\text{oc} = \sigma(\gap(\text{noc}(\vect{x}_i \circ (1-\psi(\vect{A}_i^r))))$, $p^\text{noc} = \sigma(\gap(\text{noc}(\vect{x}_i \circ (1 - \psi(\vect{A}_i^r) > \delta_\text{noc}))))$, and $\delta_\text{noc}$ is a confidence threshold.

By refining the \textbf{discriminator} $noc$ to match the masked image to the label vector $\vect{y}_i$, in which $\vect{y}_i^r = 1$, we expect it to gradually shift its attention towards secondary (and yet discriminative) regions, and, thus, to provide more useful regularization to the training of the generator.
Concomitantly, we expect the \textbf{generator} $f$ to not forget the class discriminative regions learned so far, considering (a) its learning rate is linearly decaying towards $0$; and (b) the degeneration of the masks would result in an increase of $\mathcal{L}_\text{cse}$.

\begin{figure*}[!t]
  \centering
  \includegraphics[width=\linewidth]{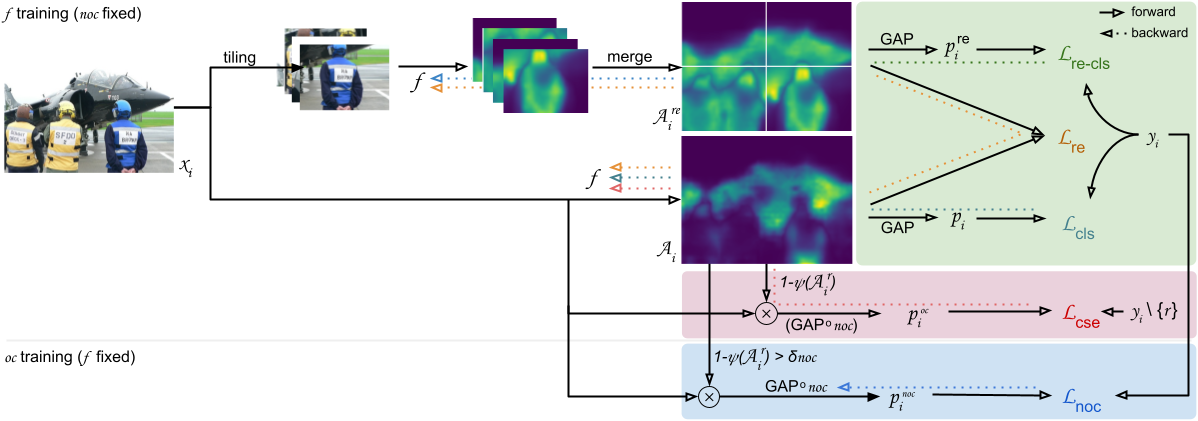}
  \caption{Detailing of all training objectives in P-NOC. In each training step, a sample $x_i$ is presented to $f$, which is optimized to produce attention maps considering the regularization provided by both Puzzle module and a ``not-so-ordinary'' classifier $noc$. $f$ is then fixed and $noc$ is refined to shift its attention to secondary discriminative regions, currently neglected by both networks.}
  \label{fig:diagram-p-noc}
\end{figure*}

Finally, to retain the properties discussed in~\cref{sec:method:poc}, we employ P-OC as our \textbf{generator}. For briefness, we refer to the combination of the P-OC and NOC-CSE strategies as P-NOC for the remaining of this work.
\cref{fig:diagram-p-noc} illustrates all elements and objectives confined in our strategy, and a pseudocode of P-NOC is provided in~\cref{alg:pnoc}.

\begin{algorithm}[!htb]
\small\caption{Proposed P-NOC algorithm}
\label{alg:pnoc}
\begin{algorithmic}[1]
\Require{Training set $\mathcal{D} = \{(\vect{x}_i, \vect{y}_i)\}$,
CAM generating networks $f$ and $noc$,
$k_\text{noc}\in\mathbb{N}$, $\delta_\text{noc}\in (0, 1)$}
\Statex
\While{not done}
\State {Sample a batch $(\vect{x}_i, \vect{y}_i)$ from $\mathcal{D}$, and $r$ from the set of classes such that $\vect{y}_i^r = 1$}
\State {\textit{// Fix $noc$ and train $f$}}
\State {Compute $\vect{A}_i=f(\vect{x}_i), and \vect{A}_i^\text{re}=\text{merge}(f(\text{tile}(\vect{x}_i)))$}
\State {Compute $\mathcal{L}_\text{P-OC}$ loss from \cref{eq:poc-loss}}
\State {Update weights of $f$ by $\nabla \mathcal{L}_\text{P-OC}$}
\If {$i\mod k_\text{noc} = 0$}
    \State {\textit{// Fix $f$ and train $noc$}}
    \State {$\hat{\vect{x}}_i = \vect{x}_i \circ (1 - \psi(\vect{A}^r_i) > \delta_\text{noc})$}
    \State {Compute $\mathcal{L}_{noc}$ from \cref{eq:pnoc:noc}}
    \State {Update weights of \textit{noc} by $\nabla\mathcal{L}_\text{noc}$}
\EndIf
\EndWhile
\end{algorithmic}
\end{algorithm}

\section{Learning Saliency Information from Weakly Supervised Data}
\label{sec:method:wssd}


We propose to utilize \textit{saliency hints}, extracted from models trained in the weakly supervised scheme, to further enrich the training of C²AM.
Our approach is inspired by recently obtained results in the task of Semi-Supervised Semantic Segmentation, in which a teacher network is used to regularize the training of a student network~\cite{wang9879387semiunreliable}.

\subsection{Method}
For every image $\vect{x}_i$, saliency hints $\hat{\vect{y}}_i$ are extracted by
\textit{thresholding} the maximum pixel activation in its CAM by a foreground constant ($\hat{\vect{y}}_i=\max_c\psi(\vect{A}_i^c) > \delta_\text{fg}$), resulting in a map that highlights regions that likely contain salient objects.
Given the often observed lack of \textit{completeness} in CAMs, only regions associated with a high activation intensity are considered as \textit{fg} hints, and thus used to reinforce a strong output classification value for the disentangling function $\varphi$.
Conversely, regions associated with a low activation intensity are discarded.

\begin{figure*}[!t]
  \centering
  \includegraphics[width=\linewidth]{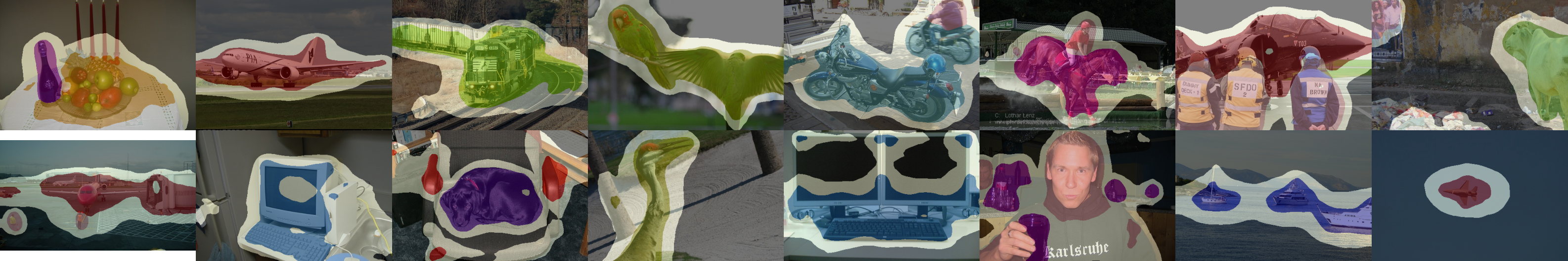}
  \caption{Example of hints (colored regions) used to train C²AM-H. Hints were extracted from segmentation priors produced by a P-NOC model, considering regions where activation intensity were lower than $\delta_\text{bg}=0.1$ and higher than $\delta_\text{fg}=0.4$.}\label{fig:cams-conf-rs269-0.1-0.4}
\end{figure*}

\cref{fig:cams-conf-rs269-0.1-0.4} illustrates examples of background and foreground hints extracted from CAMs obtained from the RS269 P-NOC model. We notice \textit{fg} hints tend to be significantly accurate (rarely activating over background regions).
Conversely, background hints are less accurate, often highlighting salient objects due to the lack of \textit{prediction completeness}.

We define C²AM-H as an extension of C²AM, in which \textit{fg} hints are employed to guide training towards a solution in which salient regions are associated with high prediction values from $\varphi$ (anchored), and all salient objects are contained within the same partition.
In practice, this implies in the addition of a new objective in the training loss described in~\cref{eq:ccam-loss}: the cross-entropy loss term between the hints $\hat{\vect{y}}_i$, and the posterior probability $p_i^{hw}$ predicted by $\varphi$. That is, C²AM-H is trained with the following multi-objective loss:
\begin{equation}
\label{eq:ccamfgh-loss}
\begin{split}
  \mathcal{L}^\mathcal{B}_\text{C²AM-H} &= \mathcal{L}^\mathcal{B}_\text{pos-f} + \mathcal{L}^\mathcal{B}_\text{pos-b} + \mathcal{L}^\mathcal{B}_\text{neg}
  + \mathcal{L}^\mathcal{B}_\text{hint} \\
    &= \frac{1}{n(n-1)}\sum_i^n\sum_j^n \mathds{1}_{[i\ne j]}(w^f_{ij}\log s^f_{ij})
    + \frac{1}{n(n-1)}\sum_i^n\sum_j^n \mathds{1}_{[i\ne j]}(w^b_{ij}\log s^b_{ij}) \\
    &- \frac{1}{n^2}\sum_i\sum_j\log (1-s^\text{neg}_{ij})
    + \lambda_{h}\sum_{i\in b}\sum_{h,w} \mathds{1}_{[A_i^{hw} > \delta_\text{fg}]}\ell_\text{cls}(\hat{y}^{hw}_i, p^{hw}_i)    
\end{split}
\end{equation}
\noindent where $\mathds{1}_{[A_i^{hw} > \delta_\text{fg}]}$ is a binary mask applied to ensure only regions associated with a normalized activation intensity greater than $\delta_\text{fg}$ are considered as foreground hints.

\subsection{Properties of C²AM-H}
\paragraph{Saliency Anchor} Contrary to C²AM --- which establishes a bi-partition of the visual receptive field without specifying which partition contains salient objects ---, C²AM-H invariably associates similar pixels, representative of salient objects (relative to the dataset of interest), to high values in $P^{hw}$.
\paragraph{Class Set Integrity} C²AM-H explicitly reinforces for the creation a bi-partition that aggregates all salient classes in the set. On the other hand, C²AM
simply projects similar regions together, implying in the risk of salient objects of different classes to be placed in different partitions. For example, in extreme scenarios, where two classes never appear spatially close to each other, or indirectly through an intermediate class.
\paragraph{Inductive Bias} C²AM considers similarity based on low and high level features from a pretrained model, which may differ from the semantic boundaries of classes in the problem at hand.
It stands to reason that additional inductive bias, related to the segmentation of the task at hand, may prove itself beneficial in the learning of a better bi-partition function by C²AM-H.

\section{Leveraging Saliency Maps in Pixel Affinity Inference}
\label{sec:method:sal_aff_labels}

\begin{figure}[!htb]
    \centering
    \includegraphics[width=\linewidth]{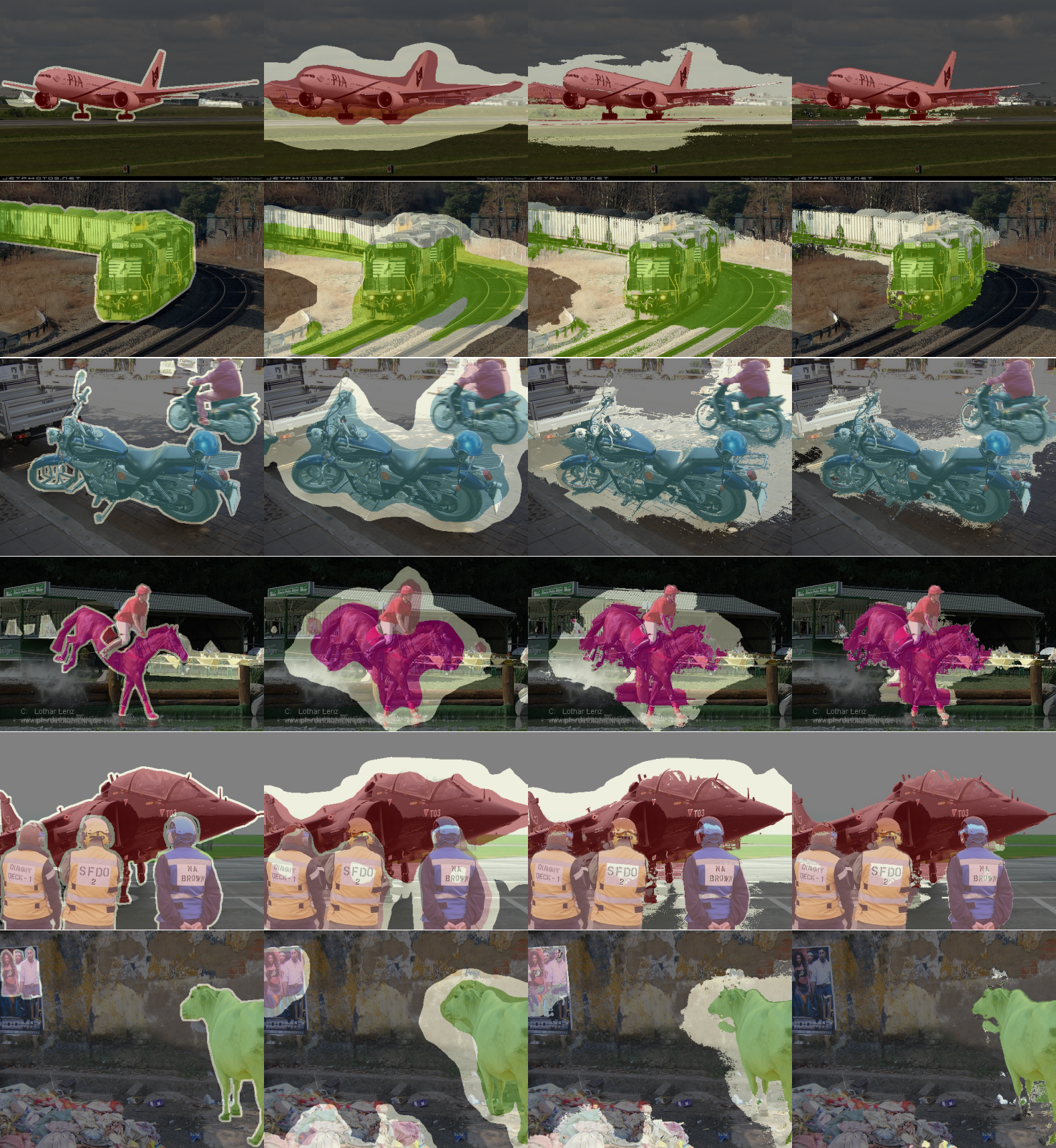}
    \caption{Comparison between the different affinity maps obtained from RS269 trained with P-NOC. From left to right: (a) images and ground-truth segmentation; (b) coarse affinity labels from priors; (c) conventional affinity labels refined with dCRF~\cite{ahn2018learning}; and (d) (our) affinity labels, obtained using both C²AM-H and dCRF.}
    \label{fig:aff-rs269-comparison-c}
\end{figure}

Affinity maps are originally devised by employing two thresholds ($\delta_{bg}$ and $\delta_{fg}$) over the prior maps to determine core regions (depicting probable \textit{bg} or \textit{fg} regions). CRF~\cite{krahenbuhl2011efficient} is used to refine both maps, and the pixels whose labels both maps cannot agree upon are marked as \textit{uncertain} and ignored during training~\cite{ahn2018learning}. Affinity maps are exemplified in the third column of \cref{fig:aff-rs269-comparison-c}.

We propose a slight modification to this procedure to incorporates the saliency maps obtained from C²AM-H: we assign (i) the background label to pixels whose saliency value are lower than a small threshold $\delta_\text{bg}$; (ii) a class label for pixels with a strong CAM signal (higher than a large threshold $\delta_\text{fg}$); and (iii) the \textit{uncertain} label for pixels with strong saliency value and weak CAM signal.

Affinity maps devised from both segmentation priors and saliency proposals are less prone to the problem of lack of \textit{prediction completeness} and are more aware of semantic contours, resulting in (a) less false negative pixels (\textit{fg} regions labels as \textit{bg}), (b) fewer uncertain regions, and (c) higher faithfulness to semantic boundaries.
We further remark that this strategy yields more affinity pairs when training the pixel-wise affinity function~\cite{ahn2018learning}, as well as more useful, ``harder'' examples that are closer to the semantic boundaries of the objects.
This property is best illustrated in the examples shown in the last column of~\cref{fig:aff-rs269-comparison-c}.

\section{Experimental Setup and Details}
\label{sec:setup}

In this section, we detail the training and evaluation procedures, as well as its hyperparameters, employed in this work.

\subsection{Obtaining Robust Semantic Segmentation Priors}
\label{sec:setup:priors}

\paragraph{Hyperparameters}
When training Puzzle-CAM, OC-CSE and P-OC strategies, we maintain all hyperparameters as originally proposed in literature: the coefficients $\lambda_\text{re}$ increases linearly from $0$ to $4$ during the first half epochs, while $\lambda_\text{cse}$ increases linearly from $0.3$ to $1$ throughout training.

For P-NOC, the foreground threshold $\delta_\text{noc}$ is set to $0.2$ following previous work. Moreover, we limit the update rate of the weights of $noc$ by employing a scheduling factor $\lambda_\text{noc}$ that linearly increases throughout training. Conjointly with the decreasing learning rate, the effective change rate of $noc$ starts and ends at zero, reaching its peak value mid-training. These settings constraint $noc$ to change more significantly in the intermediate epochs, and prevent the corruption of its internalized knowledge during (i) early and (ii) later stages of training, when (i) $f$ has not yet properly learned the primary regions of class, and (ii) to prevent $noc$ from forgetting primary features and learning incorrect ones later on.

\paragraph{Training}
Following previous work~\cite{ahn2018learning,occse9711138,puzzle9506058}, CAM-generating models are trained over the Pascal VOC 2012 dataset~\cite{everingham2015pascal} with Stochastic Gradient Descent (SGD) for $15$ epochs with linearly decaying learning rates of $0.1$ and $0.01$ (for randomly initialized weights and pretrained weights, respectively), and $1\text{e-4}$ weight decay.
We set a batch size of 32 for all architectures but ResNeSt-269, due to hardware limitations. The latter is trained with a batch of 16 samples in the Puzzle-CAM, P-OC and P-NOC training setups, and gradients are accumulated on each odd step, and only applied on even steps.
For the MS COCO 2014 dataset~\cite{lin2014microsoft}, models are trained with learning rates $0.05$ and $0.005$ for randomly initialized weights and pretrained weights, respectively.

During Pascal VOC 2012 training, image samples were resized by a random factor ranging between 0.5 to 2.0, with a patch of 512 per 512 pixels² being subsequently extracted from a random position. For the MS COCO 2014 dataset, models are trained with patches of 640 per 640 pixels².
Rand\-Augment (RA)~\cite{cubuk2020randaugment} is utilized when training vanilla models, which will later be employed as Ordinary Classifiers (OCs), to induce a higher classification effectiveness and robustness against data noise.
We only employ weak image augmentation procedures (changes in image brightness, contrast, and hue) when training Puzzle, P-OC and P-NOC.
Label Smoothing~\cite{muller2019labelsmoothing,szegedy2016rethinking} is applied to the P-OC and P-NOC strategies. 

\paragraph{Inference}

Test-Time Augmentation (TTA) is employed to produce the final segmentation priors: images are resized according to the factors 0.5, 1.0, 1.5 and 2.0, and are sequentially fed (along with their horizontal reflection) to the model, producing 8 different CAMs, which are subsequently averaged to create the segmentation prior.

\paragraph{Evaluation}
The segmentation priors are evaluated with respect to their fidelity to human-annotated semantic segmentation masks, measured through the \textit{mean Intersection over Union} (mIoU) metric.
For greater statistical stability, we train and evaluate three different models for each strategy and report their averaged scores in~\cref{sec:results:priors}.

\subsection{Learning Saliency Information from Weakly Supervised Data}
\label{sec:setup:ccamh}

C²AM and C²AM-H are trained with the same hyperparameters as Xie et al.~\cite{xie2022c2am}, except for the batch size, which is set to 32 for the ResNeSt269 architecture due to hardware limitations.
For C²AM-H, we set $\delta_{fg}=0.4$ and $\lambda_{h}=1$ after inspecting samples for each class (as illustrated in~\cref{fig:cams-conf-rs269-0.1-0.4}), and confirming a low false positive rate for \textit{fg} regions in the limited inspection subset.
We leave the search for their values as future work.

\paragraph{Training} Following Xie et al.~\cite{xie2022c2am}, we employ PoolNet~\cite{liu2019simple} as our fully-supervised saliency detector and train it over the pseudo-saliency masks obtained by C²AM-H. Its training and inference procedures are kept unchanged.

\paragraph{Evaluation} We evaluate the pseudo-saliency masks, produced by C²AM-H, with respect to their contribution towards the overall segmentation effectiveness.
This is performed by concatenating the pseudo-saliency maps of an input image with its CAM, resulting in a pseudo-semantic segmentation map that can be evaluated with common segmentation metrics (e.g., mIoU).

\subsection{Semantic Segmentation}
\label{sec:setup:segm}

To verify the effectiveness of our strategy against methods in the literature, we train a semantic segmentation model over the pseudo-semantic segmentation masks devised from P-NOC and refined with C²AM-H and \textit{random walk}.

For comparison fairness, we employ DeepLabV2~\cite{chen2017deeplab} and DeepLabV3+~\cite{chen2018encoder} architectures as semantic segmentation networks, and keep 
their original training procedures unchanged:
For DeepLabV2, we employ the ResNet-101 architecture as backbone, with weights pretrained over the ImageNet dataset.
We set a batch size of 10 and crop each training image to the size of 321 per 321 pixels².
We train the model for 10,000 iterations over the Pascal VOC 2012 dataset, using an initial learning rate of 2.5e-4.
When training the segmentation model over the MS COCO 2014 dataset, we employ 20,000 training iterations over patches of 481 per 481 pixels², using an initial learning rate of 2e-4.
For DeepLabV3+, we employ the ResNeSt-269 as backbone, and train the model for 50 epochs. The batch size is set to 16, and the initial learning rates of 0.007 and 0.004 are used for Pascal VOC 2012 and MS COCO 2014, respectively.

\section{Results}
\label{sec:results}

In this section, we report and evaluate the effectiveness of each of the aforementioned strategies.

\subsection{Evaluating Semantic Segmentation Priors}
\label{sec:results:priors}

\cref{tbl:overall-comp-miou-per-epoch} shows the mIoU achieved by various techniques, measured at the end of each training epoch.
During training, samples are resized and cropped to the common training frame sizes for performance purposes, while Test-Time Augmentation (TTA) is not employed. Hence, the intermediate measurements are estimations of the true scores (represented in the column E15).

\begin{table*}[htb]
\small\centering
\caption{The mIoU (\%) values measured in each epoch over Pascal VOC 2012 \textit{training} set, considering the architectures ResNeSt101 (RS101) and ResNeSt269 (RS269); and training strategies CAM, Puzzle (P), P-OC, and P-NOC. Scores averaged among three runs.}
\label{tbl:overall-comp-miou-per-epoch}
\bgroup
\setlength{\tabcolsep}{0.85mm}
\begin{tabular}{@{}lrrrrrrrrrrrrrrrr@{}}
\toprule
\textbf{Strategy} & {\textbf{E1}} & {\textbf{E2}} & {\textbf{E3}} & {\textbf{E4}} & {\textbf{E5}} & {\textbf{E6}} & {\textbf{E7}} & {\textbf{E8}} & {\textbf{E9}} & {\textbf{E10}} & {\textbf{E11}} & {\textbf{E12}} & {\textbf{E13}} & {\textbf{E14}} & {\textbf{E15}} & {\textbf{TTA}} \\
\midrule
RS101 CAM & 48.7 & 48.2 & \bftab{50.5} & 50.4 & 49.1 & 49.5 & 49.2 & 49.8 & 41.0 & 48.9 & 48.7 & 49.4 & 48.9 & {49.2} & 49.2 &  54.8 \\
RS269 CAM & 47.3 & 49.0 & \bftab{49.3} & 49.2 & 49.2 & 48.8 & {48.7} & 48.7 & 48.6 & 48.7 & 48.7 & 48.2 & 48.0 & 48.1 & 48.1 &  53.9 \\
\hdashline
RS101 P & 50.4 & 51.4 & 53.2 & 53.4 & 52.5 & 52.9 & 54.0 & 54.7 & 54.2 & 51.8 & 53.8 & 54.7 & 54.2 & 54.6 & \bftab{54.9} & 59.4 \\
RS269 P & 50.4 & 52.5 & 54.3 & 53.9 & 55.5 & 55.9 & 55.4 & 56.1 & 56.3 & \bftab{57.0} & 55.6 & 56.7 & 55.2 & 56.7 & 56.2 & 60.9 \\ \hdashline
RS101 P-OC & 49.6 & 50.3 & 51.5 & 51.8 & 52.5 & 51.5 & 49.0 & 49.9 & 53.2 & {52.5} & 53.4 & 54.2 & 54.9 & {55.5} & \bftab{56.0} & 59.1 \\
RS269 P-OC & 49.0 & 51.1 & 52.6 & 53.6 & 54.1 & 53.8 & 51.9 & 54.8 & 55.2 & 55.7 & 54.1 & 55.6 & 57.0 & 57.0 & \bftab{57.4} & 61.4 \\
RS269 P-OC\tiny{+LS} & 50.6 & 52.5 & 53.5 & 54.3 & 53.9 & 55.0 & 55.2 & 55.3 & 56.4 & 56.1 & 55.8 & 56.2 & 55.9 & 57.5 & \bftab{58.5} & 61.8 \\ \hdashline
RS269 P-NOC & 48.4 & 51.8 & 53.0 & 53.7 & 54.3 & 55.1 & 56.4 & 53.0 & 56.3 & 57.3 & 55.8 & 57.0 & 56.6 & 57.1 & \bftab{57.4} & 62.9 \\
RS269 P-NOC{\tiny+LS} & 49.0 & 51.6 & 53.0 & 53.9 & 55.5 & 55.5 & 55.8 & 56.4 & 57.2 & 58.3 & 56.9 & 58.9 & 58.8 & 58.9 & \bftab{59.2} & \bftab{63.6} \\
\bottomrule
\end{tabular}
\egroup
\end{table*}

Vanilla CAM and Puzzle (P) present saturation on early epochs, and a significant deterioration in mIoU for the following ones.
On the other hand, Combining Puzzle and OC-CSE (P-OC) induces a notable increase in mIoU, with a score peak on latter epochs.
On average, P-OC obtains $61.4\%$ mIoU when TTA is used, $0.6$ percent point lower ($62.0 \to 61.4\%$) than the original Puzzle, but 0.55 p.p higher ($60.9\% \to 61.4\%$) than P.
Applying \textit{label smoothing} to P-OC (P-OC{\tiny+LS}) improves TTA score by 0.33 p.p. ($61.4 \to 61.8\%$), while the adversarial training of $noc$ (P-NOC) improves mIoU by 1.5 p.p. ($61.4\% \to 62.9\%$).
Finally, combining label smoothing with P-NOC (P-NOC{\tiny+LS}) results in the best solution, with $63.6\%$ mIoU and a 1.8 p.p.~improvement ($61.8 \to 63.6\%$).

\begin{figure*}[htb]
     \centering
     \begin{subfigure}[b]{0.495\linewidth}
         \centering
         \includegraphics[width=\linewidth]{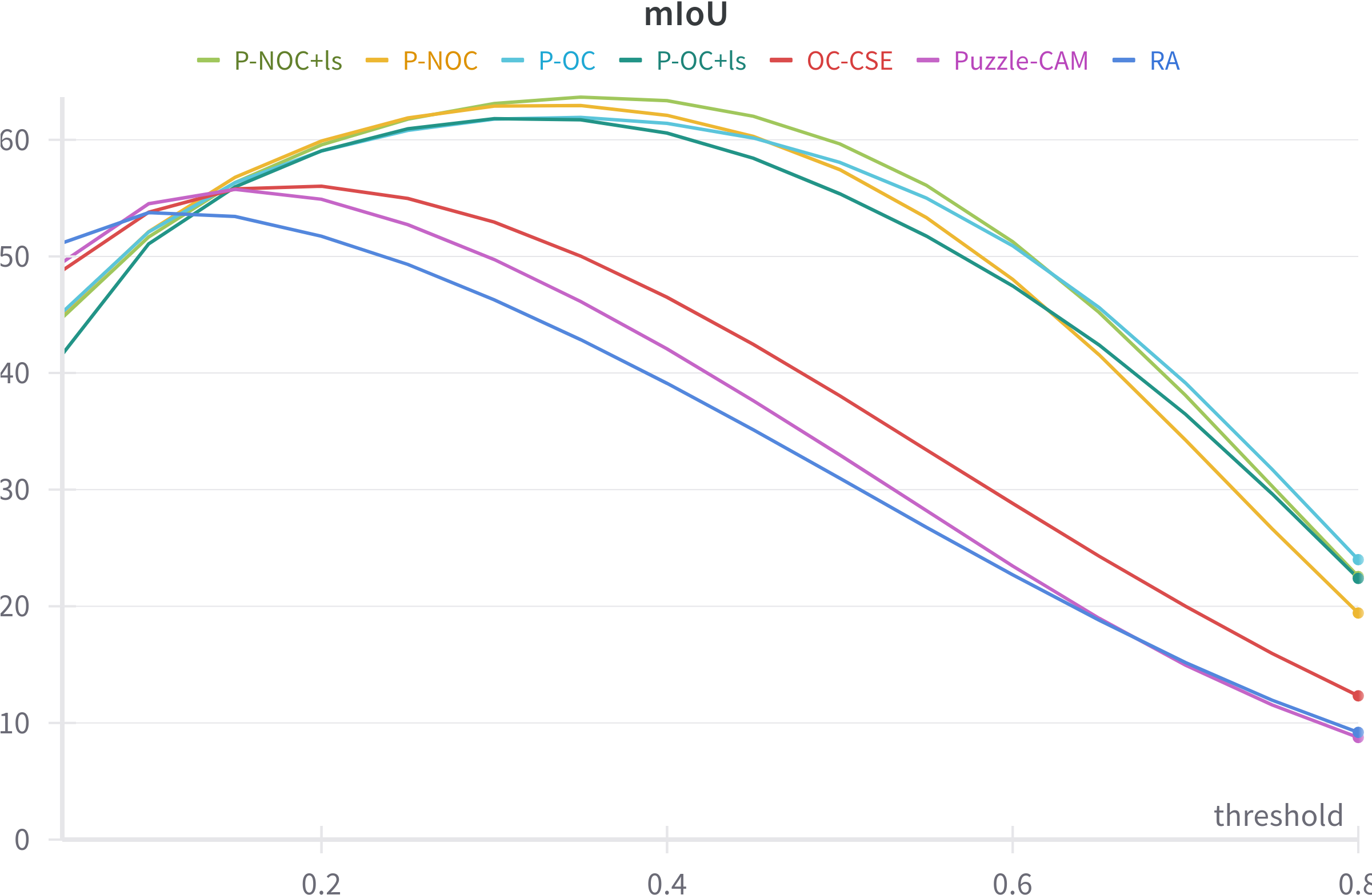}
         \caption{}
         \label{fig:comp-t-miou}
     \end{subfigure}
     \hfill
     \begin{subfigure}[b]{0.495\linewidth}
         \centering
         \includegraphics[width=\linewidth]{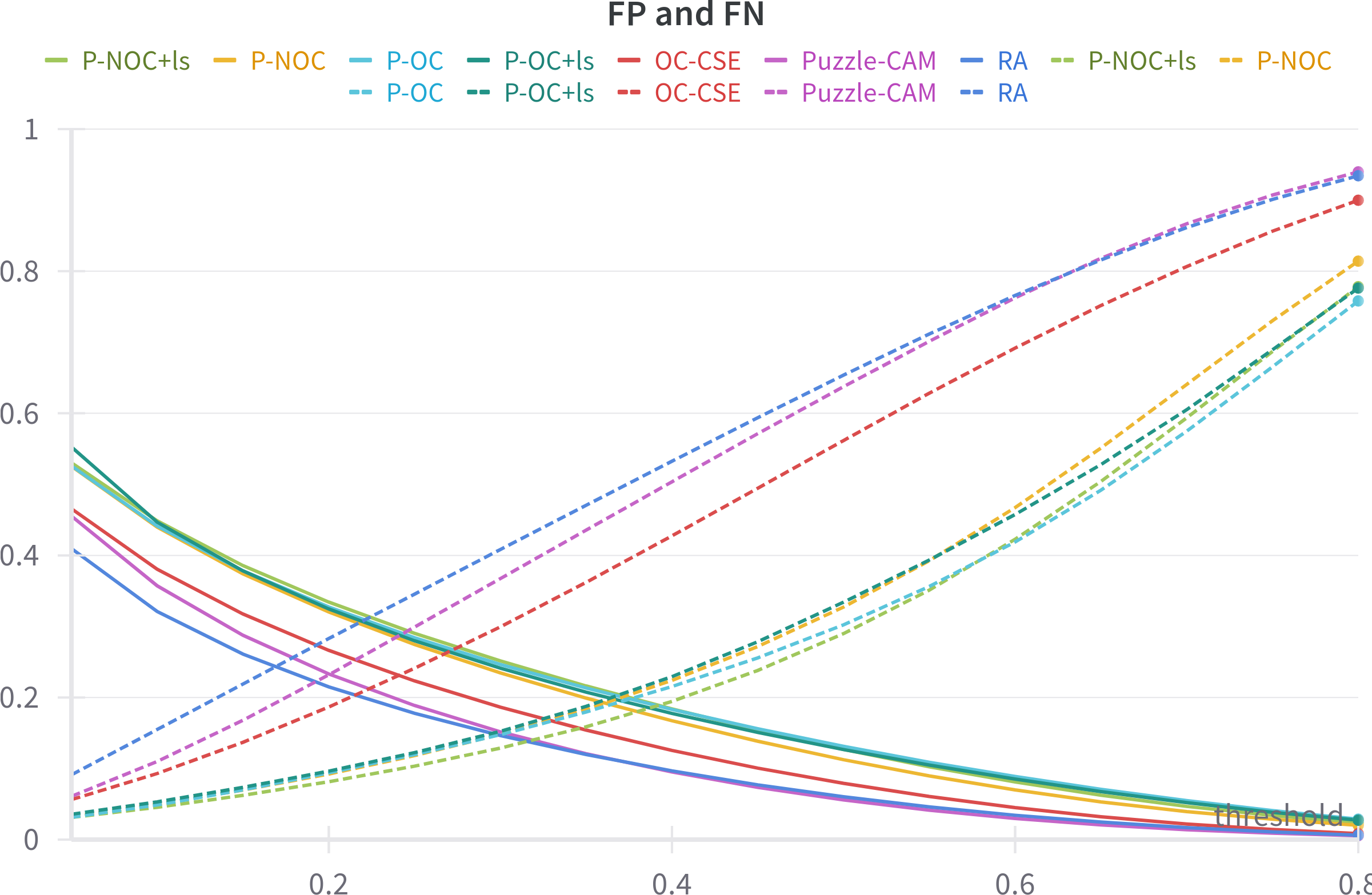}
         \caption{}
          \label{fig:comp-t-fp-fn}
     \end{subfigure}
     
     \caption{Variation in (a) mIoU and (b) False Positive/False Negative Rates when the threshold $\delta_\text{fg}$ is changed, measured over Pascal VOC 2012 \textit{training} set.}
     \label{fig:comp-t}
\end{figure*}

\cref{fig:comp-t} displays the sensitivity of mIoU and FPR/FNR to variations of in the foreground threshold. Puzzle is marginally better than the baseline (RA), while P-OC and P-NOC display higher area under the curve (mIoU) for most choices thereof. This indicates that P-OC and P-NOC not only produce higher-quality segmentation priors, but are also more robust to variations in $\delta_\text{fg}$.

\cref{tbl:poc-detailed} displays the IoU scores for each class in the Pascal VOC 2012 training set, as well as for class subgroups devised from statistical properties over these same classes.
P-NOC obtains the highest overall mIoU ($63.7\%$), as well as the highest scores for \textit{small} objects ($53.1\%$), \textit{large} objects ($69.7\%$), and for class \textit{person} and class groups \textit{animal}, \textit{room} and \textit{traffic}.
As these groups correlate with classes associated with highest observed average label set cardinality (L), we conclude that P-NOC is more indicated when inferring segmentation priors for samples associated with many classes at once.
Our approach also presents a high mIoU for the remaining groups (specially \textit{low freq.} and \textit{singleton}).

\begin{table*}[!t]
\centering
\caption{Scores (in mIoU) observed for class groups in Pascal VOC 2012 \textit{train} set, considering segmentation priors generated by CAM, OC-CSE (CSE), Puzzle-CAM (P), P-OC{\tiny+LS} (POC), and P-NOC{\tiny+LS} (PNOC). Class-specific properties, such as label occurrence (O), average relative size (\%S), class co-occurrence rate (\%C) and average of label set cardinality (L) were averaged and listed for inspection purposes.}
\label{tbl:poc-detailed}
\small
\setlength{\tabcolsep}{1mm}
\begin{tabular}{@{}llllrrrrrrrrr@{}}
\toprule
\textbf{Class} & \textbf{Group} & \textbf{Size} & {\textbf{Freq.}} & \textbf{O} & \textbf{\%S} & {\textbf{\%C}} & \textbf{L} & \textbf{CAM} & \textbf{CSE} & \textbf{P} & \textbf{POC} & \textbf{PNOC} \\
\midrule
background & - & - & - & - & 69.5\% & - & - & 81.0 & 82.2 & 86.0 & 85.6 & \bftab{86.8} \\
a.plane & singleton & small & mid & 88 & 11.8\% & 9.1\% & 1.1 & 47.5 & 53.6 & 61.4 & \bftab{62.3} & 58.2 \\
bike & traffic & small & low & 65 & 6.4\% & 76.9\% & 2.2 & 32.2 & 34.7 & 38.6 & \bftab{45.0} & 42.8 \\
bird & singleton & small & high & 105 & 11.8\% & 11.4\% & 1.2 & 49.5 & 51.5 & 71.4 & 62.8 & \bftab{71.8} \\
boat & singleton & small & low & 78 & 10.8\% & 32.1\% & 1.4 & 40.8 & 35.4 & \bftab{51.3} & 43.9 & 49.5 \\
bottle & room & small & mid & 87 & 9.5\% & 70.1\% & 2.3 & 49.0 & 60.8 & 56.0 & \bftab{65.9} & 64.5 \\
bus & traffic & large & low & 78 & 31.5\% & 51.3\% & 1.7 & 72.1 & 65.9 & 78.4 & 75.1 & \bftab{80.6} \\
car & traffic & mid & high & 128 & 15.5\% & 61.7\% & 1.9 & 62.6 & 63.3 & 70.4 & \bftab{74.6} & 74.4 \\
cat & singleton & large & high & 131 & 28.6\% & 26.0\% & 1.3 & 54.8 & 78.4 & \bftab{83.7} & 79.8 & 83.6 \\
chair & room & small & high & 148 & 10.6\% & 87.8\% & 2.4 & 30.7 & 24.0 & 27.2 & 23.1 & \bftab{28.9} \\
cow & singleton & mid & low & 64 & 18.0\% & 29.7\% & 1.4 & 55.1 & 68.2 & \bftab{73.6} & 70.6 & 69.0 \\
d.table & room & large & mid & 82 & 22.5\% & 95.1\% & 2.6 & \bftab{52.5} & 46.3 & 39.2 & 51.0 & 52.2 \\
dog & animal & large & high & 121 & 19.8\% & 38.0\% & 1.5 & 61.3 & 61.2 & \bftab{80.8} & 76.9 & 79.5 \\
horse & animal & large & low & 68 & 19.1\% & 47.1\% & 1.6 & 55.9 & 68.9 & 69.0 & 69.8 & \bftab{70.3} \\
m.bike & traffic & large & mid & 81 & 19.6\% & 56.8\% & 1.7 & 67.8 & 68.3 & 73.2 & \bftab{76.6} & 75.9 \\
person & person & mid & high & 442 & 15.2\% & 83.0\% & 2.1 & 63.6 & 63.8 & 50.3 & \bftab{70.2} & \bftab{70.2} \\
p.plant & room & small & mid & 82 & 11.2\% & 63.4\% & 2.1 & 46.8 & 42.5 & 56.3 & \bftab{57.2} & 56.1 \\
sheep & singleton & large & low & 63 & 19.7\% & 19.0\% & 1.3 & 55.3 & 75.0 & \bftab{75.9} & 73.9 & 73.6 \\
sofa & room & large & mid & 93 & 21.6\% & 80.6\% & 2.4 & \bftab{50.0} & 45.3 & 35.5 & 34.8 & 43.6 \\
train & singleton & large & mid & 83 & 26.6\% & 20.5\% & 1.2 & 63.9 & 45.5 & \bftab{68.2} & 50.9 & 68.0 \\
tv & room & mid & mid & 84 & 15.5\% & 65.1\% & 2.1 & 38.6 & 41.5 & 33.8 & \bftab{49.8} & 37.2 \\
\rowcolor{gray!10}
{overall $\mu$} & \multicolumn{3}{l}{\textit{all}} & 108.6 & 19.8\% & 51.2\% & 1.8 & 53.9 & 56.0 & 61.0 & 61.9 & \bftab{63.7} \\
{overall $\sigma$} & \multicolumn{3}{l}{\textit{all}} & 82.0 & 6.7\% & 26.8\% & 0.5 & \bftab{11.2} & 14.9 & 17.4 & 15.7 & 15.6 \\
\midrule
\multicolumn{4}{l}{\textbf{Size}} \\
small & \multicolumn{3}{l}{a.plane bike bird boat} & 93.3 & 10.3\% & 50.1\% & 1.8 & 42.4 & 43.2 & 51.7 & 51.5 & \bftab{53.1} \\
      & \multicolumn{3}{l}{bottle chair p.plant} \\
mid & \multicolumn{3}{l}{car cow person tv} & 179.5 & 16.1\% & 59.9\% & 1.9 & 55.0 & 59.2 & 57.0 & \bftab{66.3} & 62.7 \\
large & \multicolumn{3}{l}{bus cat d.table dog horse} & 88.9 & 23.2\% & 48.3\% & 1.7 & 59.3 & 61.6 & 67.1 & 65.4 & \bftab{69.7} \\
      & \multicolumn{3}{l}{m.bike sheep sofa train} \\
\midrule
\multicolumn{4}{l}{\textbf{Frequency}} \\
low freq. & \multicolumn{3}{l}{bike boat bus cow horse sheep} & 69.3 & 17.6\% & 42.7\% & 1.6 & 51.9 & 58.0 & \bftab{64.5} & 63.0 & 64.3 \\
mid freq. & \multicolumn{3}{l}{a.plane bottle d.table m.bike} & 85.0 & 17.3\% & 57.6\% & 1.9 & 52.0 & 50.5 & 52.9 & 56.1 & \bftab{57.0} \\
          & \multicolumn{3}{l}{p.plant sofa train tv} \\
high freq. & \multicolumn{3}{l}{bird car cat chair dog person} & 179.2 & 16.9\% & 51.3\% & 1.7 & 53.8 & 57.0 & 64.0 & 64.6 & \bftab{68.1} \\
\midrule
\multicolumn{4}{l}{\textbf{Group}} \\
singleton & \multicolumn{3}{l}{a.plane bird boat sheep train} & 87.4 & 18.2\% & 21.1\% & 1.3 & 52.4 & 58.2 & \bftab{69.3} & 63.5 & 67.7 \\
person & \multicolumn{3}{l}{person} & 442.0 & 15.2\% & 83.0\% & 2.1 & 63.6 & 63.8 & 50.3 & \bftab{70.2} & \bftab{70.2} \\
animal & \multicolumn{3}{l}{dog horse} & 94.5 & 19.5\% & 42.5\% & 1.5 & 58.6 & 65.0 & \bftab{74.9} & 73.3 & \bftab{74.9} \\
room & \multicolumn{3}{l}{bottle chair d.table} & 96.0 & 15.2\% & 77.0\% & 2.3 & 44.6 & 43.4 & 41.3 & 47.0 & \bftab{47.1} \\
     & \multicolumn{3}{l}{p.plant sofa tv} \\
traffic & \multicolumn{3}{l}{bike bus car m.bike} & 88.0 & 18.3\% & 61.7\% & 1.9 & 58.7 & 58.0 & 65.1 & 67.8 & \bftab{68.4} \\
\midrule
\multicolumn{4}{l}{\textbf{Correlation}} \\
{$\rho$ O} &  &  &  & 100.0 & -6.5 &  30.0 &  20.4 & \bftab{19.9} &  10.4 &         -12.8 & 10.7 & 10.7 \\
$\rho$ \%S &  &  &  & -6.5 & 100.0 & -18.0 & -23.4 & \bftab{70.3} &  51.5 &          48.4 & 38.3 & 53.1 \\
$\rho$ \%C &  &  &  & 30.0 & -18.0 & 100.0 &  97.6 &        -22.9 & -38.4 & \bftab{-71.0} & -42.9 & -51.4 \\
$\rho$ L   &  &  &  & 20.4 & -23.4 &  97.6 & 100.0 &        -32.9 & -44.5 & \bftab{-75.8} & -49.7 & -58.2 \\
\bottomrule
\end{tabular}
\end{table*}

P-OC achieves the second-best overall mIoU ($61.9\%$), with the highest scores for classes \textit{person} and \textit{tv}, and consistently high scores for the remaining groups.
On the other hand, Puzzle-CAM scores third overall ($61.0\%$), achieving the best score for infrequent classes (\textit{low freq.}) ($64.5\%$) and \textit{singletons} ($69.3\%$), while also displaying notable effectiveness for \textit{large} objects ($67.1\%$).
Conversely, it presents low mIoU for class \textit{person} ($50.3\%$), and objects in the \textit{room} class group ($41.3\%$), often associated with images contain many objects of different classes. %
Lastly, segmentation priors obtained from vanilla CAM result in the best mIoU scores for classes \textit{dining table} and \textit{sofa}, as well as high score for class group \textit{room}, notwithstanding presenting the worst overall mIoU ($53.9\%$).

\begin{table*}[!t]
\small\centering
\caption{The mIoU (\%) scores observed after each epoch over the MS COCO 2014 \textit{training} set; and the mIoU scores observed over the \textit{validation} set when TTA is employed. Training strategies considered are: CAM (CAM), Puzzle (P), P-OC, and P-NOC. %
}\label{tab:results-priors-mscoco}
\bgroup
\setlength{\tabcolsep}{1mm}
\begin{tabular}{@{}lrrrrrrrrrrrrrrrr@{}}
\toprule
  \textbf{Strategy} & {\textbf{E1}} & {\textbf{E2}} & {\textbf{E3}} & {\textbf{E4}} & {\textbf{E5}} & {\textbf{E6}} & {\textbf{E7}} & {\textbf{E8}} & {\textbf{E9}} & {\textbf{E10}} & {\textbf{E11}} & {\textbf{E12}} & {\textbf{E13}} & {\textbf{E14}} & {\textbf{E15}} & \textbf{TTA} \\
  \midrule
  RS269 CAM & 28.2 & 29.4 & 29.8 & 30.0 & 30.3 & 30.2 & 30.7 & \bftab{30.4} & 30.3 & 30.3 & 30.3 & 30.3 & 30.3 & 30.2 & 30.2  & {33.7} \\
  \hdashline
  RS269 P & 34.9 & 36.6 & 38.0 & \bftab{38.3} & 37.6 & 36.5 & 33.1 & 35.0 & 34.3 & 33.5 & 29.9 & 30.9 & 31.7 & 30.9 & 30.7 & {36.8} \\
  \hdashline
  RS269 P-OC & 32.2 & 34.5 & 36.1 & 36.8 & 37.3 & 37.3 & 36.5 & 36.8 & 36.5 & 35.7 & 36.1 & \bftab{36.6} & 35.9 & 36.3 & 36.3 & {37.3} \\
  \hdashline
  RS269 P-NOC & 32.7 & 34.3 & 35.3 & 35.4 & 36.3 & 36.3 & 36.9 & 37.1 & 37.5 & 38.2 & 37.7 & 38.4 & \bftab{38.4} & 38.2 & 38.2 & \bftab{40.7} \\
  \bottomrule
\end{tabular}
\egroup
\end{table*}

\begin{table*}[!t]
\centering
\caption{Scores evaluated (in IoU) over the MS COCO 2014 \textit{validation} set, for segmentation priors generated by various WSSS methods.}
\label{tbl:coco-detailed}
\small
\begin{tabular}{@{}lrrrr|lrrrr@{}}
\toprule
\textbf{Class} & \textbf{CAM} & \textbf{P} & \textbf{P-OC} & \textbf{P-NOC} & \textbf{Class} & \textbf{CAM} & \textbf{P} & \textbf{P-OC} & \textbf{P-NOC} \\
\midrule
background & 74.1 & 37.7 & 37.3 & \bftab{78.9} & wine glass & 29.8 & 19.9 & 29.3 & \bftab{37.3} \\
person & 37.2 & 17.0 & 17.2 & \bftab{50.0}     & cup & 28.0 & 30.0 & 29.9 & \bftab{32.1} \\
bicycle & 44.7 & 40.9 & \bftab{50.8} & 48.9    & fork & 12.8 & 14.0 & 15.5 & \bftab{18.8} \\
car & 35.2 & 31.0 & 34.1 & \bftab{42.1}        & knife & 13.7 & 11.5 & 10.9 & \bftab{21.2} \\
motorcycle & 60.6 & 52.3 & 62.5 & \bftab{67.4} & spoon & 10.4 & 7.6 & 7.7 & \bftab{14.4} \\
airplane & 47.2 & 47.4 & 47.6 & \bftab{52.3}   & bowl & \bftab{22.0} & 22.0 & 21.3 & 18.5 \\
bus & 51.2 & 41.9 & 49.0 & \bftab{69.0}        & banana & 47.6 & 59.0 & 60.3 & \bftab{60.6} \\
train & 53.6 & 47.6 & 47.5 & \bftab{59.2}      & apple & 37.3 & 41.6 & 42.7 & \bftab{46.7} \\
truck & 40.0 & 27.1 & 32.9 & \bftab{45.6}      & sandwich & 35.3 & 42.2 & \bftab{42.3} & 40.7 \\
boat & 34.1 & 32.9 & 33.2 & 38.0               & orange & 48.0 & 52.7 & 54.6 & \bftab{56.0} \\
traffic light & 18.8 & \bftab{34.3} & 30.3 & 26.0 & broccoli & 42.8 & 36.5 & 46.3 & \bftab{49.8} \\
fire hydrant & 46.5 & \bftab{59.8} & 52.5 & 57.4  & carrot & 30.4 & 21.4 & 24.0 & \bftab{35.1} \\
stop sign & 40.5 & \bftab{59.6} & 47.7 & 54.0     & hot dog & 38.5 & 51.3 & \bftab{51.4} & 50.1 \\
parking meter & 45.9 & \bftab{60.5} & 58.8 & 56.3 & pizza & 40.6 & 59.7 & \bftab{60.3} & 58.5 \\
bench & 34.2 & 33.4 & 32.7 & 39.9                 & donut & 40.3 & \bftab{57.4} & 54.2 & 53.6 \\
bird & 36.7 & \bftab{50.3} & 44.7 & 47.6          & cake & 39.5 & 44.0 & 43.4 & \bftab{45.7} \\
cat & 43.4 & 66.5 & 65.1 & \bftab{67.2}           & chair & \bftab{30.1} & 18.0 & 18.7 & 25.9 \\
dog & 49.7 & \bftab{63.7} & 62.4 & 63.7           & couch & \bftab{37.5} & 27.9 & 27.1 & 31.9 \\
horse & 44.9 & 55.1 & 56.2 & \bftab{60.1}         & potted plant & 31.7 & 24.8 & 26.3 & \bftab{32.9} \\
sheep & 44.3 & 60.0 & 57.5 & \bftab{63.1}         & bed & 39.1 & 37.2 & 36.6 & \bftab{40.5} \\
cow & 43.5 & 60.2 & 58.1 & \bftab{64.8}           & dining table & 18.7 & 18.1 & \bftab{19.4} & 14.8 \\
elephant & 49.3 & 70.1 & 68.4 & \bftab{73.2}      & toilet & 45.1 & 54.4 & 54.4 & \bftab{56.0} \\
bear & 40.1 & 73.1 & 70.5 & \bftab{73.6}          & tv & 33.5 & 32.0 & 34.1 & \bftab{38.7} \\
zebra & 57.1 & 67.6 & 63.7 & \bftab{71.4}         & laptop & 42.8 & 38.9 & 41.5 & \bftab{47.6} \\
giraffe & 50.1 & 59.9 & 58.8 & \bftab{64.8}       & mouse & \bftab{21.8} & 12.1 & 12.4 & 18.0 \\
backpack & 25.8 & 19.8 & 18.8 & \bftab{27.0}      & remote & 23.8 & 48.2 & \bftab{48.4} & 22.0 \\
umbrella & 43.1 & 30.0 & 38.9 & \bftab{51.6}      & keyboard & 46.6 & 55.7 & \bftab{56.5} & 56.2 \\
handbag & \bftab{19.1} & 10.7 & 9.6 & 12.5        & cell phone & 36.8 & 49.0 & 48.6 & \bftab{49.4} \\
tie & 18.7 & 11.0 & \bftab{20.6} & 15.8           & microwave & \bftab{39.0} & 33.6 & 34.3 & 38.3 \\
suitcase & 43.5 & 39.5 & 38.1 & \bftab{43.6}      & oven & \bftab{32.3} & 25.8 & 26.6 & 29.9 \\
frisbee & 21.2 & \bftab{50.4} & 48.9 & 23.1       & toaster & 25.2 & 19.5 & 19.1 & \bftab{26.2} \\
skis & 6.6 & 6.2 & \bftab{11.3} & 8.3             & sink & 25.1 & 21.4 & 20.5 & \bftab{29.3} \\
snowboard & 19.7 & 13.1 & 11.8 & 28.6             & refrigerator & \bftab{32.4} & 20.6 & 19.9 & 27.9 \\
sports ball & 9.9 & 35.6 & \bftab{42.4} & 22.5    & book & 29.9 & 30.6 & 30.9 & \bftab{35.8} \\
kite & 18.0 & 22.0 & 24.4 & \bftab{31.0}          & clock & 29.6 & \bftab{46.9} & 26.9 & 24.4 \\
baseball bat & 8.8 & 19.7 & \bftab{23.8} & 15.9   & vase & \bftab{30.7} & 26.5 & 25.8 & 29.5 \\
baseball glove & 5.2 & 14.7 & \bftab{18.2} & 9.5  & scissors & 34.9 & \bftab{49.2} & 47.8 & 45.9 \\
skateboard & 15.0 & 19.8 & \bftab{22.8} & 18.1    & teddy bear & 43.0 & 63.9 & 63.3 & \bftab{64.0} \\
surfboard & 21.0 & 20.4 & 17.4 & \bftab{27.8}     & hair drier & 11.7 & 11.3 & 10.4 & \bftab{21.3} \\
tennis racket & 20.4 & 26.2 & \bftab{36.1} & 18.3 & toothbrush & 20.9 & 20.6 & 21.1 & \bftab{27.7} \\ 
bottle & 28.8 & 33.7 & 33.3 & \bftab{35.6}        & \cellcolor{gray!10} overall & \cellcolor{gray!10} 35.3 & \cellcolor{gray!10} 39.6 & \cellcolor{gray!10} 40.4 & \cellcolor{gray!10} \bftab{44.5} \\
\bottomrule
\end{tabular}
\end{table*}

\cref{tab:results-priors-mscoco} displays the mIoU scores for the segmentation priors over the MS COCO 2014 dataset.
Employing P-NOC to train the CAM-generating network entails substantial improvement of mIoU compared to both Puzzle and P-OC: $3.9$ p.p. ($36.3\% \to 38.2\%$) and $3.4$ p.p. ($37.3\% \to 40.7\%$), respectively.

Similarly, \cref{tbl:coco-detailed} displays IoU scores for each class in the MS COCO 2014 dataset. Once again, we observe that P-NOC achieves the best mIoU among all methods.

\clearpage
\subsection{Evaluating Pseudo-Saliency Information Learned}
\label{sec:results:wssd}

\begin{figure*}[htb]
    \centering
    \includegraphics[width=\linewidth]{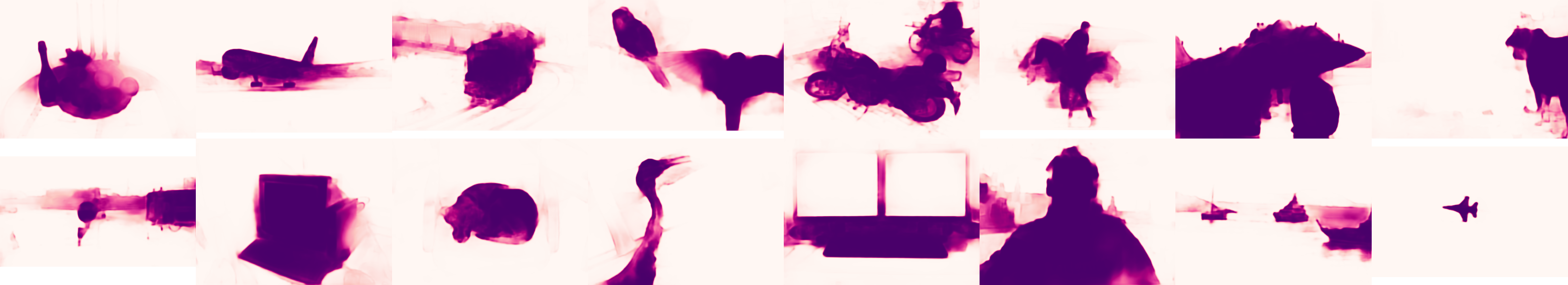}
    \caption{Saliency maps from a simple saliency detector (PoolNet) trained over priors devised by C²AM-H.}\label{fig:sal-ccamfgh-rs269}
\end{figure*}

\cref{fig:sal-ccamfgh-rs269} illustrates examples of saliency masks obtained by a simple saliency detection network, when trained over pseudo-masks devised from {C²AM-H}.
We observe better fidelity to semantic boundaries in these examples, compared to the unrefined segmentation priors, as well as better separation between salient and non-salient visual element (e.g., airplanes and trains are detected while railroad and landing site are suppressed).

In conformity with literature~\cite{xie2022c2am}, we evaluate the effective refinement resulted from combining saliency maps devised by C²AM (baseline) and C²AM-H (trained with additional \textit{fg} hints) with semantic segmentation priors, and report the observed mIoU scores over the Pascal VOC 2012 dataset in \cref{tbl:ccam-ablation}.

\begin{table}[htb]
\centering
\caption{The ablation study for C²AM-H, considering various backbone architectures (B.bone) and segmentation priors (Segm. Priors). Scores for both saliency priors (P.) and saliency maps refined with PoolNet (R.) are reported in mIoU (\%) over Pascal VOC 2012 \textit{train} set.}
\label{tbl:ccam-ablation}
\bgroup
\begin{tabular}{@{}llllrr@{}}
\toprule
\textbf{Method} & \textbf{B.bone} & \textbf{Hints} & \textbf{Segm. Priors} & \textbf{P.} & \textbf{R.} \\
\midrule
C²AM~\cite{xie2022c2am} & RN50  & -           & RN50  P~\cite{puzzle9506058} & 56.6  & 65.5     \\\hdashline
C²AM~\cite{xie2022c2am} & RN50  & -           & RS269 P & 59.1  & 65.3     \\
C²AM~\cite{xie2022c2am} & RN50  & -           & RS269 P-OC   & 60.8  & \bftab{67.3}     \\
C²AM~\cite{xie2022c2am} & RN50  & -           & RS269 P-OC{\tiny+LS} & \bftab{61.2}  & 67.2 \\
\hdashline
C²AM-H   & RS269  & RS269 P-OC & RS269 P-OC   & 66.8  & 68.6     \\
C²AM-H   & RS269  & RS269 P-OC & RS269 P-OC{\tiny+LS} & 67.3  & 68.8 \\
C²AM-H   & RS269  & RS269 P-OC{\tiny+LS} & RS269 P-OC{\tiny+LS} & 67.3 & 69.2 \\
C²AM-H   & RS269  & RS269 P-OC{\tiny+LS} & RS269 P-NOC{\tiny+LS} & 67.7 & 70.6 \\
C²AM-H   & RS269  & RS269 P-NOC{\tiny+LS} & RS269 P-NOC{\tiny+LS} & \bftab{67.9} & \bftab{70.8} \\
\midrule
C²AM~\cite{xie2022c2am} & RN50   & -          & GT           & 63.4  & 65.0     \\
\hdashline
C²AM-H   & RN50   & RS269 P-OC & GT           & 64.8  & -        \\
C²AM-H   & RS101  & RS269 P-OC & GT           & 69.6  & -        \\
C²AM-H   & RS269  & RS269 P-OC & GT           & 69.9  & 70.9     \\
C²AM-H   & RS269  & RS269 P-OC{\tiny{+LS}}  & GT & 70.3 & 71.7 \\
C²AM-H   & RS269  & RS269 P-NOC{\tiny{+LS}} & GT & \bftab{70.6} & \bftab{72.4} \\
\bottomrule
\end{tabular}
\egroup
\end{table}

Firstly, we observe that the employment of priors devised from P-OC (4th and 5th rows) substantially improves mIoU when compared to the baseline ($56.6\%\to 61.2\%$) and RS269 trained with Puzzle~($59.1\%\to 61.2\%$).
Moreover, we observe consistent score improvement when employing hints of foreground regions in C²AM-H training (6th to 10th rows), compared to its unsupervised counterpart.
Our best strategy (priors and hints from P-NOC\textsubscript{+LS}) scores 70.8\% mIoU after refinement with PoolNet, 1.6 p.p.~higher ($69.2\%\to 70.8\%$) than the second-best strategy (priors and hints from P-OC\textsubscript{+LS}), and 5.3 p.p.~higher ($65.5\% \to 70.8\%$) than the baseline.


To provide a more comprehensive and fair evaluation, we isolated the contribution of the saliency maps to the score by employing the ground-truth annotations (GT) as segmentation priors (last six rows in~\cref{tbl:ccam-ablation}).
That is, the prediction for a given pixel in the image is considered correct if that pixel is predicted as \textit{salient} by the detector, and it is annotated with any class $c$ other than the background. Conversely, a pixel annotated with $c$ and predicted as non-salient (or annotated as \textit{bg} and predicted as salient) is considered a \textit{miss}.
In this evaluation setup, the baseline (C²AM RN50) scored 65.0\% mIoU, while our best strategy (C²AM-H, using hints from RS269 P-NOC{\tiny{+LS}}) achieves 72.4\% mIoU (a 7.4 p.p.~increase).

\begin{figure*}[htb]
    \centering
    \includegraphics[width=\linewidth]{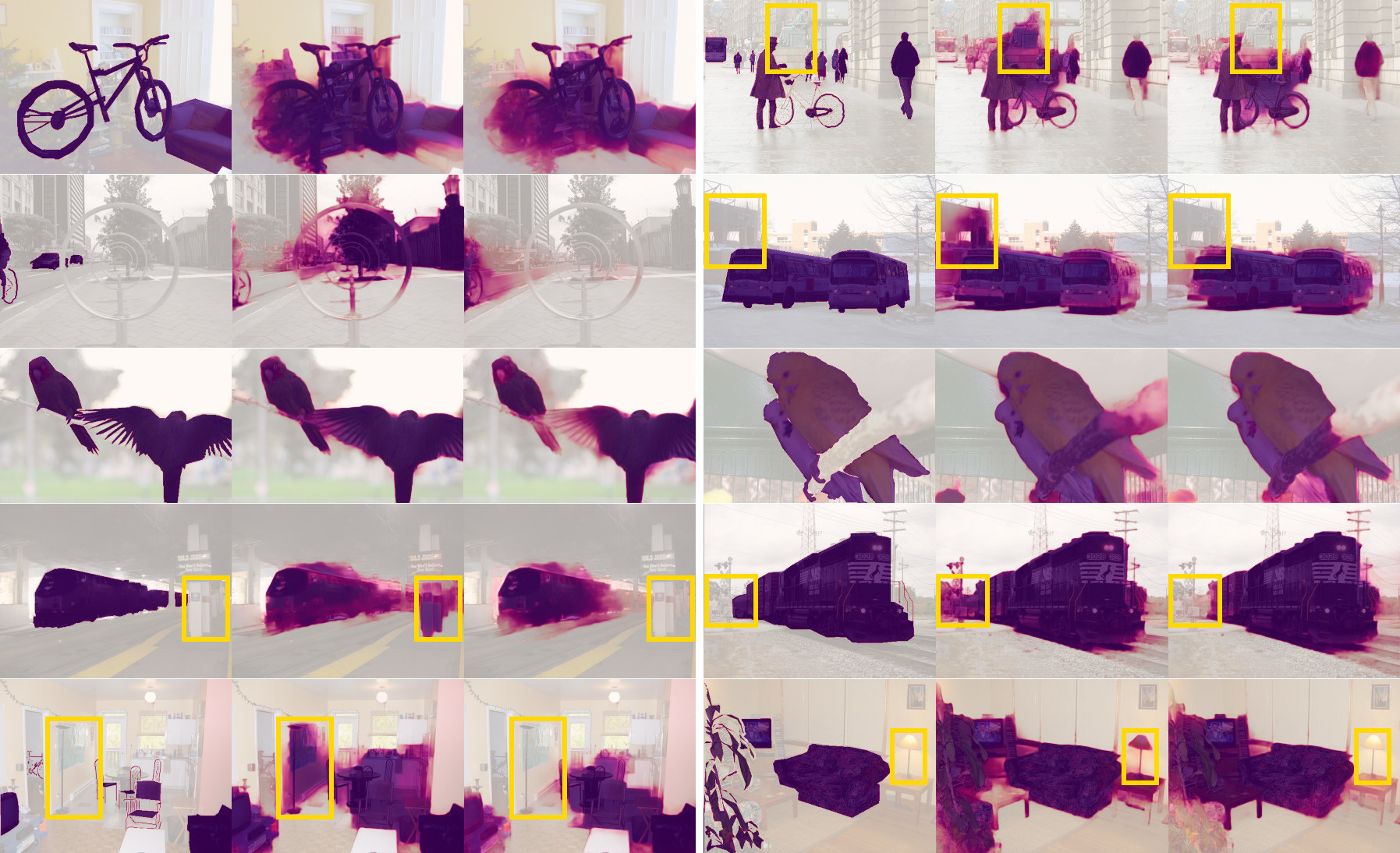}
    \caption{Comparison between saliency proposals of C²AM and C²AM-H. From left to right: ground-truth annotations; C²AM; and C²AM-H.}
    \label{fig:sal-comparison}
\end{figure*}

{\cref{fig:sal-comparison} illustrates examples of saliency proposals devised from C²AM and C²AM-H. We observe that C²AM often predicts background regions close to the objects of interest as salient.
On the other hand, C²AM-H successfully integrates the supervised information available in the image-level labels, ignoring background areas and correlated objects. }

\subsection{Evaluating Pseudo-Semantic Segmentation Masks}
\label{sec:results:rw}

\cref{tbl:all-overall} describes the mIoU scores obtained throughout the different stages of training. The utilization of saliency maps (P-OC C²AM-H) increases the mIoU of the pseudo-segmentation maps by 0.40 p.p. ($73.50\% \to 73.90\%$).
Further training of $noc$ (+NOC) improves mIoU by 1.63 p.p. ($73.90\% \to 75.53\%$) and 1.88 p.p. ($72.53\% \to 74.41\%$) over training and validation subsets, respectively.
Furthermore, pseudo-segmentation masks devised from P-NOC achieve a mIoU score of $47.74\%$ over the MS COCO 2014 validation dataset, 11.34 p.p. ($36.40\% \to 47.74$) higher than the efficacy reported by OC-CSE.

\begin{table}[!htb]
    \centering\caption{Ablation studies of pseudo segmentation masks produced by \textit{RW} over priors, measured in mIoU (\%) over Pascal VOC 2012 training (\textit{train}) and validation (\textit{val}) sets, and over MS COCO 2014 validation (\textit{val}) set.}
    \label{tbl:all-overall}
    \bgroup
    \setlength{\tabcolsep}{1mm}
    \begin{tabular}{@{}lcccrrr@{}}
    \toprule
    \textbf{Method} & \textbf{+LS} & \textbf{+C²AM-H} & \textbf{+NOC} & \multicolumn{2}{c}{\textbf{VOC12}} & \textbf{COCO14} \\
    & & & & \textbf{\textit{Train}} & \textbf{\textit{Val}} & \textbf{\textit{Val}} \\
    \midrule
    P      &            &             &            & 71.35         & 70.67 & - \\
    \hdashline
    P-OC   &            &             &            & 73.50         & 72.08 & - \\
    P-OC   & \checkmark &             &            & 71.45         & 70.15 & - \\
    P-OC   &            & \checkmark  &             & 73.90         & 72.53 & - \\
    P-OC   & \checkmark & \checkmark  &             & 73.07         & 72.14 & - \\
    P-OC   & \checkmark &             & \checkmark & 73.22         & 72.83 & - \\
    P-OC   & \checkmark & \checkmark  & \checkmark & \bftab{75.53} & \bftab{74.41} & 47.74 \\
    \bottomrule
    \end{tabular}
    \egroup
\end{table}

\subsection{Comparison with the State of the Art}
\label{sec:results:segm}

We compare our solution to other approaches in the literature in~\cref{tbl:sota-comparison}. Training a segmentation model over the pseudo-segmentation masks devised by P-NOC results in competitive mIoU scores (70.3\% and 70.9\% over validation and testing sets, respectively), even when compared to modern approaches employing complex topologies, such as Transformers, or employing additional saliency information.

\begin{table}[htb]
    \centering
    \caption{Comparison with SOTA methods on Pascal VOC 2012, measured in mIoU (\%){\protect\footnotemark}. Supervision: $\mathcal{F}$ fully-supervised; $\mathcal{I}$: image-level; $\mathcal{S}$: saliency. \textdagger pre-trained over Imagenet-21k~\cite{ridnik2021imagenet}; \textdaggerdbl pre-trained over MS COCO.}
    \label{tbl:sota-comparison}
    \setlength{\tabcolsep}{1.4mm}
    \begin{tabular}{@{}llllrr@{}}
      \toprule
      \textbf{Method} & \textbf{Sup.} & \textbf{Backbone} & \textbf{Seg.} & \textbf{\textit{Val}} & \textbf{\textit{Test}} \\
      \midrule
      DeepLabV1~\cite{chen2017deeplab} & $\mathcal{F}$ & W-ResNet38 & V1 & 78.1 & 78.2 \\
      \hdashline
      AffinityNet~\cite{ahn2018learning} & $\mathcal{I}$ & W-ResNet38 & V1 & 61.7 & 63.7 \\
      SEAM~\cite{wang2020self}           & $\mathcal{I}$ & W-ResNet38 & V1 & 64.5 & 65.7 \\
      SEAM+CONTA~\cite{zhang2020causal}  & $\mathcal{I}$ & W-ResNet38 & V1 & 66.1 & 66.7 \\
      OC-CSE~\cite{occse9711138}         & $\mathcal{I}$ & W-ResNet38 & V1 & 68.4 & 68.2 \\
      ADELE~\cite{liu2022adaptive}       & $\mathcal{I}$ & W-ResNet38 & V1 & 69.3 & 68.8 \\
      MCT-Former~\cite{xi9879800token}   & $\mathcal{I}$ & W-ResNet38 & V1 & 71.9 & \bftab{71.6} \\
      \hdashline
      ICD~\cite{fan2020learning} & $\mathcal{I}$             & ResNet-101 & V1 & 64.1 & 64.3 \\ 
      ICD~\cite{fan2020learning} & $\mathcal{I}+\mathcal{S}$ & ResNet-101 & V1 & 67.8 & 68.0 \\
      EPS~\cite{lee2021railroad} & $\mathcal{I}+\mathcal{S}$ & ResNet-101 & V1 & 71.0 & \bftab{71.8} \\ 
      \hdashline
      ToCo~\cite{ru2023token}           & $\mathcal{I}$ & ViT-B & - & 69.8 & 70.5 \\
      ToCo$^\dagger$~\cite{ru2023token} & $\mathcal{I}$ & ViT-B & - & 71.1 & \bftab{72.2} \\
      \midrule
      DeepLabV2~\cite{chen2017deeplab} & $\mathcal{F}$ & ResNet-101 & V2 & 76.8 & 76.2 \\
      \hdashline
      IRNet~\cite{ahn2019weakly}   & $\mathcal{I}$               & ResNet50 & V2 & 63.5 & 64.8 \\
      AdvCAM~\cite{lee2021anti}    & $\mathcal{I}$               & ResNet-101 & V2 & 68.1 & 68.0 \\
      RIB~\cite{lee2021reducing}   & $\mathcal{I}$               & ResNet-101 & V2 & 68.3 & 68.6 \\
      AMR~\cite{qin2022activation} & $\mathcal{I}$               & ResNet-101 & V2 & 68.8 & 69.1 \\
      AMN~\cite{lee2022threshold}  & $\mathcal{I}$               & ResNet-101 & V2 & 69.5 & 69.6 \\
      SIPE~\cite{chen2022sipe}     & $\mathcal{I}$               & ResNet-101 & V2 & 68.8 & 69.7 \\ 
      URN~\cite{li2022uncertainty} & $\mathcal{I}$               & ResNet-101 & V2 & 69.5 & 69.7 \\
      RIB~\cite{lee2021reducing}   & $\mathcal{I}+\mathcal{S}$   & ResNet-101 & V2 & 70.2 & 70.0 \\
      SGWS~\cite{yi2022sgws}       & $\mathcal{I}$               & ResNet-101 & V2 & 70.5 & 70.5 \\ 
      AMN$^\ddagger$~\cite{lee2022threshold} & $\mathcal{I}$     & ResNet-101 & V2 & 70.7 & 70.6 \\
      EPS~\cite{lee2021railroad}   & $\mathcal{I}+\mathcal{S}$   & ResNet-101 & V2 & \bftab{70.9} & 70.8 \\
      ViT-PCM~\cite{rossetti2022max} & $\mathcal{I}$             & ResNet-101 & V2 & 70.3 & 70.9 \\
      \rowcolor{gray!10}
      P-NOC{\tiny+LS+C²AM-H}$^a$ & $\mathcal{I}$ & ResNet-101 & V2 & 70.3 & \bftab{70.9} \\
      \midrule
      DeepLabV3+ & $\mathcal{F}$ & ResNeSt-269 & V3+ & 80.6 & 81.0 \\
      \hdashline
      {Puzzle-CAM~\cite{puzzle9506058}} & $\mathcal{I}$ & {ResNeSt-269} & V3+ & {71.9} & {72.2} \\
      \rowcolor{gray!10} P-OC{\tiny+C²AM-H}$^b$ & $\mathcal{I}$ & ResNeSt-269& V3+ & 71.4 & 72.4 \\
      \rowcolor{gray!10}  P-NOC{\tiny+LS+C²AM-H}$^c$ & $\mathcal{I}$ & ResNeSt-269& V3+ & \bftab{73.8} & \bftab{73.6} \\
      \bottomrule
    \end{tabular}
\end{table}

\footnotetext{
Official evaluation reports:
$^a$ \href{http://host.robots.ox.ac.uk:8080/anonymous/KYTP7D.html}{KYTP7D}, 
$^b$ \href{http://host.robots.ox.ac.uk:8080/anonymous/PAVLA8.html}{PAVLA8}, 
and $^c$ \href{http://host.robots.ox.ac.uk:8080/anonymous/GUEA7B.html}{GUEA7B}.
}

A noticeable improvement is obtained when training DeepLabV3+ model over our pseudo-masks. Proposals from P-OC{\tiny+C²AM-H} entail a solution with 71.4\% and 72.4\% mIoU scores over the VOC12 validation and testing sets, respectively, slightly outscoring Puzzle-CAM ($72.2\% \to 72.4\%$) in the test set, and the remaining approaches by a considerable margin.

Finally, training DeepLabV3+ over masks obtained from P-NOC{\tiny+LS+C²AM-H} results in our best semantic segmentation model, entailing a 1.1 p.p.~improvement over P-OC ($72.4\%\to 73.5\%$).

\begin{figure*}[htb]
    \centering
    \includegraphics[width=\linewidth]{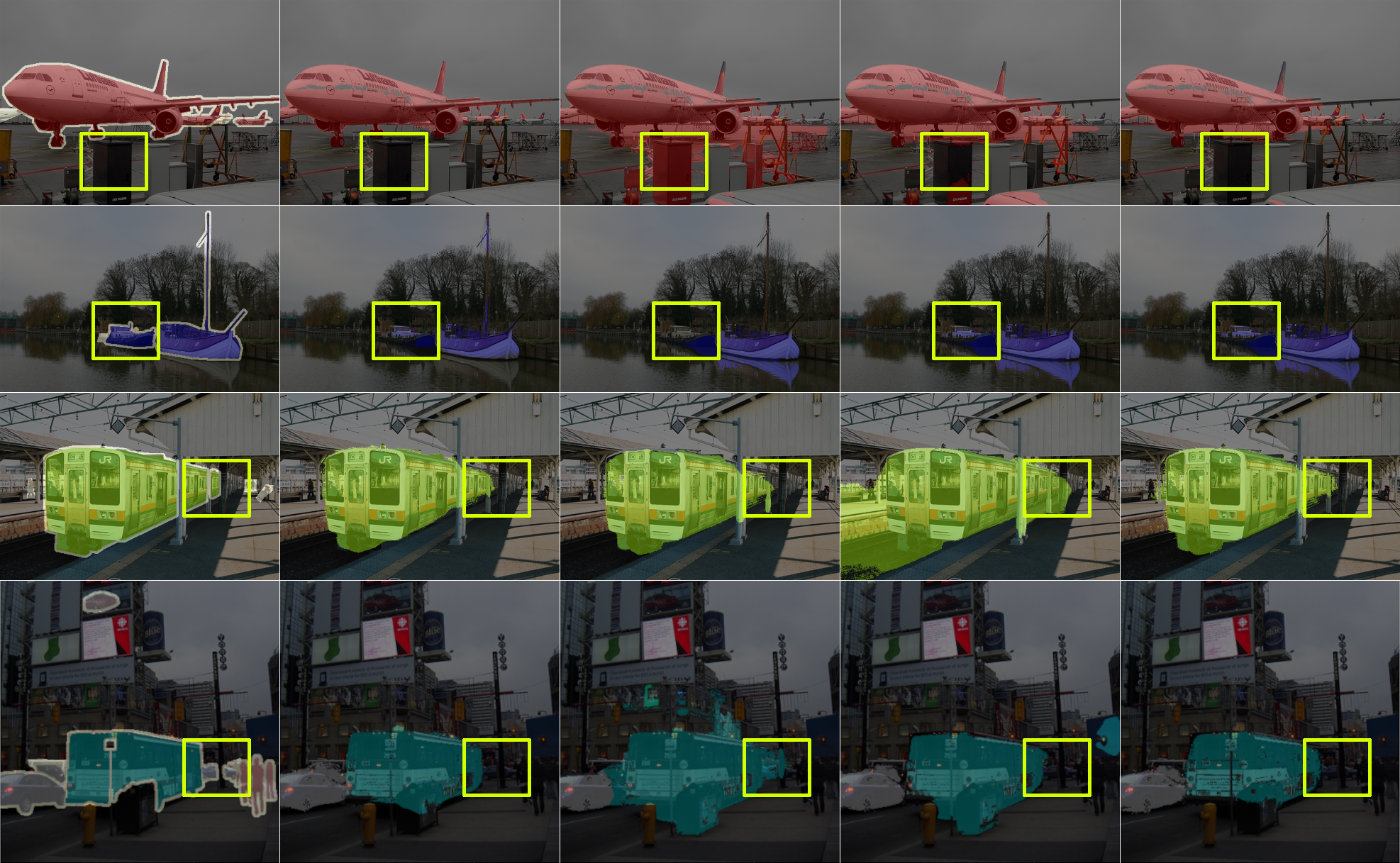}
    \caption{Comparison between predictions by DeepLabV3+ models trained over semantic segmentation masks devised by different methods. From left to right: ground-truth annotations; fully-supervised model; Puzzle-CAM; P-OC and P-NOC.}
    \label{fig:d3p-comparison-4}
\end{figure*}

\begin{figure*}[t!]
     \centering
     \begin{subfigure}[b]{\linewidth}
         \centering
         \includegraphics[width=\linewidth]{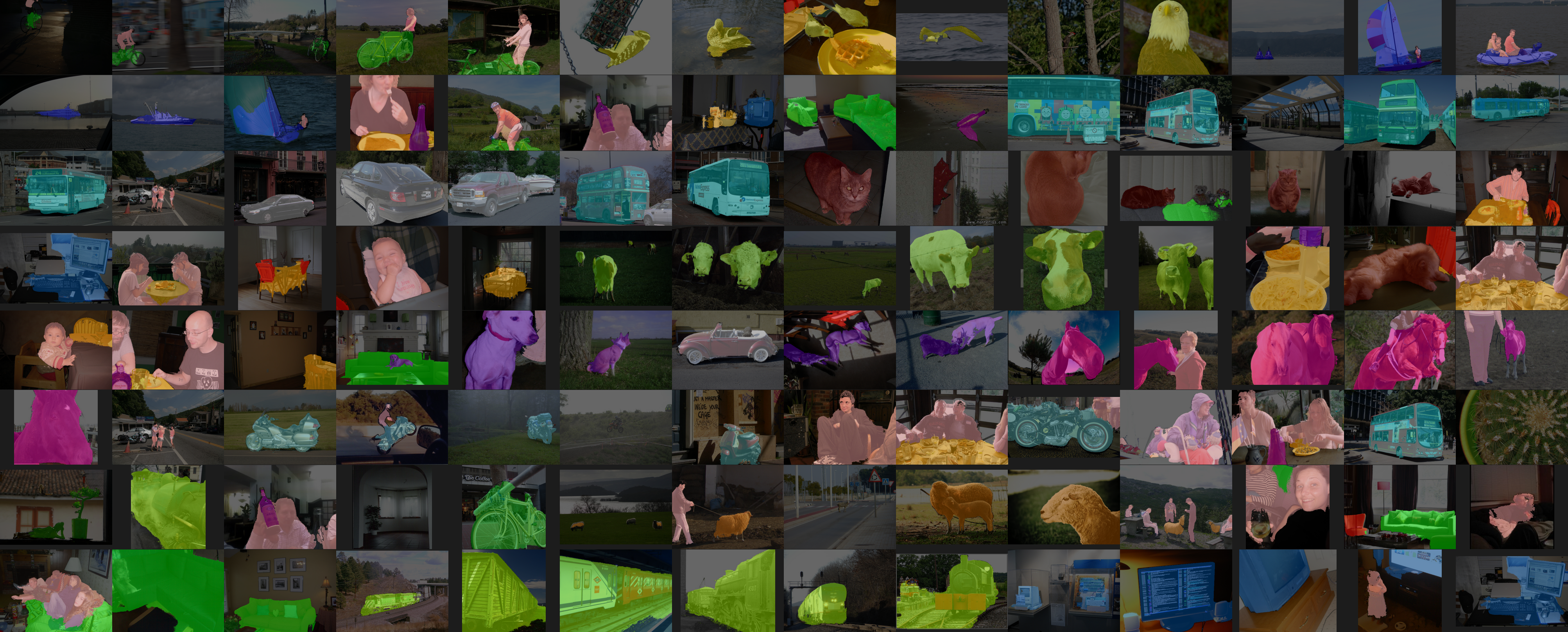}
         \caption{P-OC{\tiny+LS} Pascal VOC 2012}
         \label{fig:dlv3-fgh-poc}
     \end{subfigure}
     \hfill
     \begin{subfigure}[b]{\linewidth}
         \centering
         \includegraphics[width=\linewidth]{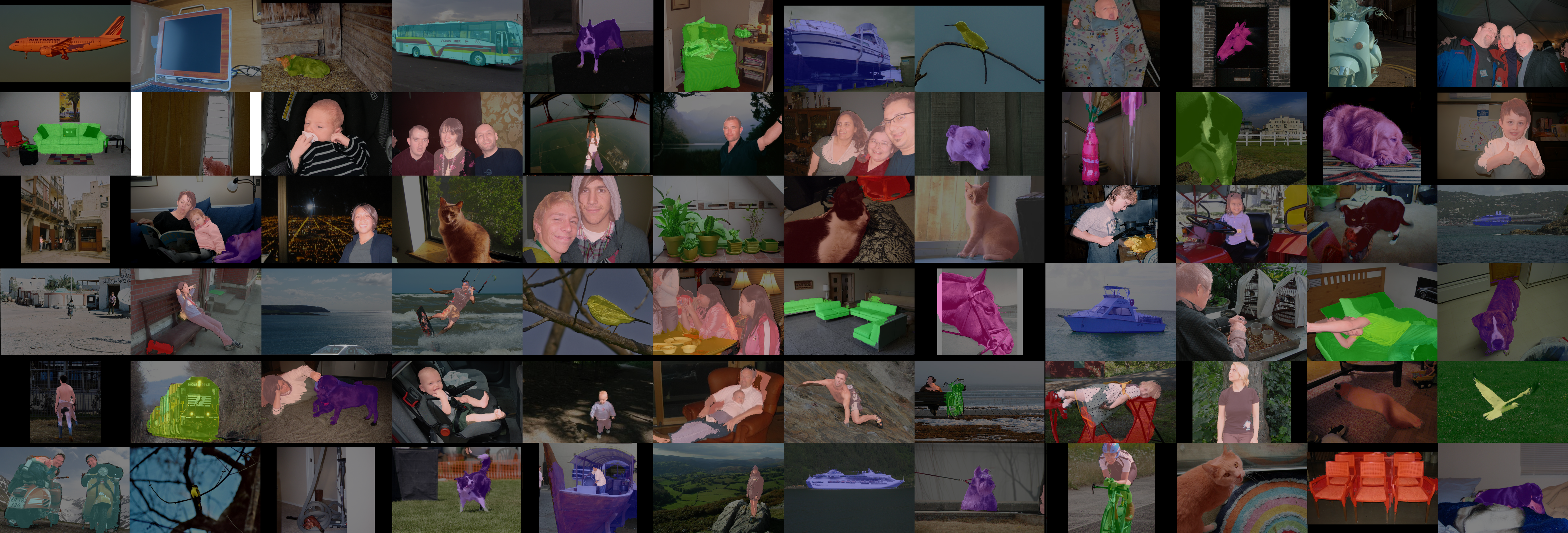}
         \caption{P-NOC{\tiny+LS} Pascal VOC 2012}
          \label{fig:dlv3-fgh-pnoc}
     \end{subfigure}
     \hfill
     \begin{subfigure}[b]{\linewidth}
         \centering
         \includegraphics[width=\linewidth]{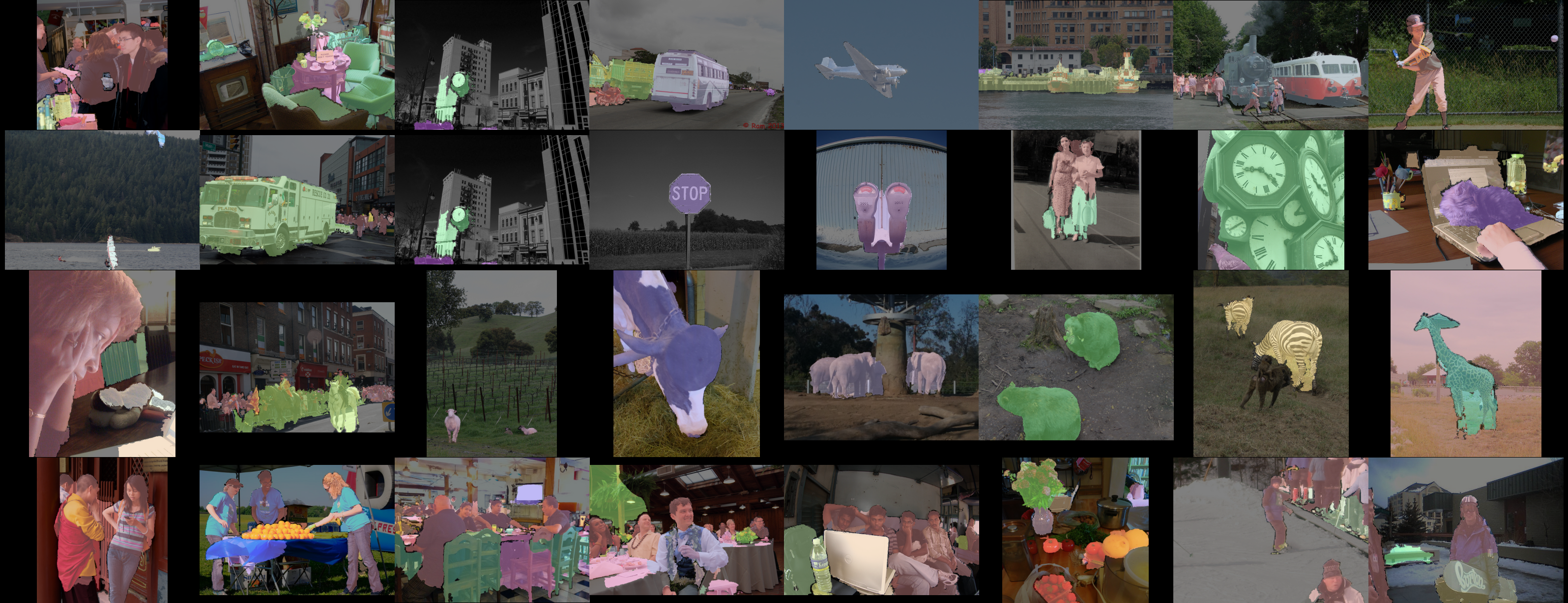}
         \caption{P-NOC+ls MS COCO 2014}
         \label{fig:coco-dlv3-fgh-pnoc}
     \end{subfigure}
        \caption{A and B: Segmentation examples for Pascal VOC 2012 by DeepLabV3+, trained over Pascal VOC 2012. Pseudo-training masks were devised from (a) P-OC and (b) P-NOC{\tiny+LS}, respectively, and refined with C²AM-H, random walk and dCRF. B: P-NOC{\tiny+LS}. C: pseudo-segmentation masks for MS COCO 2014, devised by P-NOC and refined with C²AM-H, random walk and dCRF.}
        \label{fig:dlv3-fgh-voc-coco}
\end{figure*}

\cref{fig:d3p-comparison-4} illustrates and compares semantic segmentation proposals by the DeepLabV3+ models (last four rows in \cref{tbl:sota-comparison}), trained on pseudo-semantic segmentation masks devised by Puzzle, P-OC and P-NOC.
The model trained with pseudo-masks devised from P-NOC (last column) produces the closest proposals to the fully-supervised solution (second column).
More examples of the semantic segmentation predictions over the Pascal VOC 2012 and MS COCO 2014 datasets are illustrated in~\cref{fig:dlv3-fgh-voc-coco}.

\cref{tbl:voc-test-iou} display IoU scores achieved by our approach and the best methods in the literature, providing a class-level comparison between them.

We also evaluate P-NOC over the more challenging MS COCO 2014 dataset, and present the results in~\cref{tbl:sota-comparison-coco14}.
DeepLabV2-ResNet101 Our approach achieves 42.9\% mIoU, outperforming OC-CSE and more recent WSSS methods.
Employing DeepLabV3+ entail an increase of 1.7 p.p.~over its DeepLabV2 counterpart ($42.9\% \to 44.6\%$), resulting in competitive mIoU compared to state of the art.


\begin{table}[htb]
    \centering
    \caption{Comparison with SOTA methods on MS COCO 2014 dataset, measured in mIoU (\%). \textdagger pre-trained over Imagenet-21k~\cite{ridnik2021imagenet}.}
    \label{tbl:sota-comparison-coco14}
    \setlength{\tabcolsep}{1.6mm}
    \begin{tabular}{@{}llllr@{}}
    \toprule
    \textbf{Method} & \textbf{Sup.} & \textbf{Backbone} & \textbf{Seg.} & \textbf{\textit{Val}} \\
    \midrule
    OC-CSE~\cite{occse9711138}         & $\mathcal{I}$ & W-ResNet38 & V1 & 36.4 \\
    MCT-Former~\cite{xi9879800token}   & $\mathcal{I}$ & W-ResNet38 & V1 & 42.0 \\
    \hdashline
    ToCo~\cite{ru2023token}            & $\mathcal{I}$ & ViT-B       & -   & 41.3 \\
    ToCo$^\dagger$~\cite{ru2023token}  & $\mathcal{I}$ & ViT-B       & -   & 42.3 \\
    \midrule
    IRNet~\cite{ahn2019weakly}         & $\mathcal{I}$ & ResNet-50  & V2 & 32.6 \\
    IRN+CONTA~\cite{zhang2020causal}   & $\mathcal{I}$ & ResNet-50  & V2 & 33.4 \\
    EPS~\cite{lee2021railroad}         & $\mathcal{I}+\mathcal{S}$ & VGG16 & V2 & 35.7 \\
    PPM~\cite{li2021pseudo}            & $\mathcal{I}$ & ScaleNet  & V2 & 40.2 \\
    SIPE~\cite{chen2022sipe}           & $\mathcal{I}$ & ResNet-101 & V2 & 40.6 \\
    URN~\cite{li2022uncertainty}       & $\mathcal{I}$ & ResNet-101 & V2 & 40.7 \\
    \rowcolor{gray!10}
    P-NOC{\tiny+C²AM-H}                & $\mathcal{I}$ & ResNet-101 & V2 & 42.9 \\
    RIB~\cite{lee2021reducing}         & $\mathcal{I}$ & ResNet-101 & V2 & 43.8 \\
    AMN~\cite{lee2022threshold}        & $\mathcal{I}$ & ResNet-101 & V2 & 44.7 \\
    \midrule
    \rowcolor{gray!10}
    P-NOC{\tiny+C²AM-H}                & $\mathcal{I}$ & ResNeSt-269 & V3+ & 44.6 \\
    \bottomrule
    \end{tabular}
\end{table}

\begin{landscape}
    \begin{table*}[!t]
        \small\centering
        \caption{Intersection over Union (IoU \%) obtained by segmentation models over the Pascal VOC 2012 test set, after being trained over pseudo segmentation masks devised from P-OC and P-NOC, and refined with C²AM-H and \textit{random walk}.
        }\label{tbl:voc-test-iou}
        \bgroup\def\arraystretch{1.2}
        \setlength{\tabcolsep}{1.8mm}
        \small\begin{tabular}{@{}lrrrrrrrrrrrrrrrrrrrrrr@{}}
        \toprule
        Method & bg  & plane & bike & bird & boat & bottle & bus & car & cat & chair  & cow & table & dog & horse  & mbk. & person & plant & sheep & sofa & train & tv   & avg.\\ \midrule
        \multicolumn{23}{l}{\textcolor{gray}{DeepLabV1}} \\
        AffinityNet & 89.1 & 70.6 & 31.6 & 77.2 & 42.2 & 68.9 & 79.1 & 66.5 & 74.9 & 29.6 & 68.7 & 56.1 & 82.1 & 64.8 & 78.6 & 73.5 & 50.8 & 70.7 & 47.7 & 63.9 & 51.1 & 63.7 \\
        OC-CSE & 90.2 & 82.9 & 35.1 & 86.8 & 59.4 & 70.6 & 82.5 & 78.1 & 87.4 & 30.1 & 79.4 & 45.9 & 83.1 & 83.4 & 75.7 & 73.4 & 48.1 & \bftab{89.3} & 42.7 & 60.4 & 52.3 & 68.4 \\
        MCT-Former & \bftab{92.3} & 84.4 & 37.2 & 82.8 & 60.0 & 72.8 & 78.0 & 79.0 & 89.4 & 31.7 & 84.5 & 59.1 & 85.3 & 83.8 & 79.2 & 81.0 & 53.9 & 85.3 & \bftab{60.5} & 65.7 & 57.7 & 71.6 \\
        EPS & 91.9 & \bftab{89.0} & \bftab{39.3} & 88.2 & 58.9 & 69.6 & 86.3 & 83.1 & 85.8 & 35.0 & 83.6 & 44.1 & 82.4 & \bftab{86.5} & 81.2 & 80.8 & 56.8 & 85.2 & 50.5 & \bftab{81.2} & 48.4 & 71.8 \\
        \hline
        \multicolumn{23}{l}{\textcolor{gray}{DeepLabV2}} \\
        AMN & 90.7 & 82.8 & 32.4 & 84.8 & 59.4 & 70.0 & 86.7 & 83.0 & 86.9 & 30.1 & 79.2 & 56.6 & 83.0 & 81.9 & 78.3 & 72.7 & 52.9 & 81.4 & 59.8 & 53.1 & \bftab{56.4} & 69.6 \\
        ViT-PCM & 91.1 & 88.9 & 39.0 & 87.0 & 58.8 & 69.4 & 89.4 & 85.4 & 89.9 & 30.7 & 82.6 & \bftab{62.2} & 85.7 & 83.6 & 79.7 & \bftab{81.6} & 52.1 & 82.0 & 26.5 & 80.3 & 42.4 & 70.9 \\
        \rowcolor{gray!10} P-NOC & 91.1 & 82.9 & 34.8 & 85.5 & 59.3 & 69.4 & 90.1 & 83.5 & 90.2 & 32.4 & 81.1 & 60.7 & 84.7 & 81.3 & 78.1 & 75.9 & 57.0 & 83.6 & 58.9 & 71.1 & 37.9 & 70.9 \\
        \hline
        \multicolumn{23}{l}{\textcolor{gray}{DeepLabV3+}} \\
        Puzzle-CAM & 91.1 & 87.2 & 37.4 & 86.8 & 61.5 & 71.3 & 92.2 & 86.3 & 91.8 & 28.6 & 85.1 & 64.2 & 91.9 & 82.1 & 82.6 & 70.7 & 69.4 & 87.7 & 45.5 & 67.0 & 37.8 & 72.3 \\
        \rowcolor{gray!10} P-OC & {91.6} & {86.7} & {38.3} & \bftab{89.3} & {61.1} & \bftab{74.8} & {92.0} & \bftab{86.6} & {89.9} & {20.5} & \bftab{85.8} & {57.0} & {90.2} & {83.5} & 83.4 & {80.8} & \bftab{68.0} & {87.0} & {47.1} & {62.8} & {43.1} & {72.4} \\
        \rowcolor{gray!10} P-NOC & 91.7 & 89.1 & 38.3 & 80.9 & \bftab{65.4} & 70.1 & \bftab{93.8} & 85.5 & \bftab{93.4} & \bftab{37.3} & 83.6 & 61.3 & \bftab{92.8} & 84.1 & \bftab{83.8} & 80.7 & 63.6 & 82.0 & 53.3 & 76.7 & 36.8 & \bftab{73.6} \\
        \bottomrule
        \end{tabular}
        \egroup
    \end{table*}
\end{landscape}

\clearpage

\section{Conclusions}
\label{sec:conclusions}

In this work, we investigated complementary WSSS techniques, analyzing the properties that most influenced their segmentation effectiveness. We found that \textit{prediction completeness}, sensitivity to semantic contours of objects, and robustness against noise were paramount components when performing segmentation from weakly supervised data.

We devised NOC-CSE: the adversarial training of CAM generating networks to mitigate the attention deficit to marginal features observed in the Class-Specific Adversarial framework.
When combined with complementary strategies, our approach produced competitive semantic segmentation priors, with noticeably higher mIoU than the baseline.

Subsequently, we proposed C²AM-H, an extension of C²AM that incorporates hints of salient regions to devise more robust salient maps in a weakly supervised fashion. We employed the pseudo-saliency maps learned in the creation of more robust affinity labels, entailing in a better refinement process of the aforementioned semantic segmentation priors.
Our empirical results suggest that these interventions provide noticeable improvement in all stages, resulting in solutions that are both more effective and robust against data noise.

As future work, we will (a) study the effect of strong augmentation techniques (e.g., ClassMix) to further strengthen our solution; (b) evaluate our approach over more difficult domains, such as ill-balanced datasets or functional segmentation problems, containing fuzzy visual patterns that may prove difficult to learn otherwise; and (c) devise forms to reduce computational footprint.

\section*{Acknowledgments}
The authors would like to thank CNPq (grants 140929/2021-5, 309330/2018-1, and 304836/2022-2) and LNCC/MCTI for providing HPC resources of the SDumont supercomputer.

{\small
\bibliographystyle{model1-num-names}
\bibliography{neurips_2022}
}




\end{document}